\documentclass[journal]{IEEEtran}

\usepackage{amsmath,amsfonts}
\usepackage{algorithmic}
\usepackage{array}
\usepackage[caption=false,font=normalsize,labelfont=sf,textfont=sf]{subfig}
\usepackage{textcomp}
\usepackage{stfloats}
\usepackage{url}
\usepackage{verbatim}
\usepackage{subcaption}
\usepackage{graphicx}
\usepackage{multirow}
\usepackage{cite}
\def\BibTeX{{\rm B\kern-.05em{\sc i\kern-.025em b}\kern-.08em
    T\kern-.1667em\lower.7ex\hbox{E}\kern-.125emX}}
\usepackage{academicons}
\usepackage{hyperref}
\usepackage{svg}
\usepackage{balance}
\usepackage{booktabs}
\usepackage[export]{adjustbox}
\usepackage{xcolor}
\usepackage{tikz}

\newcommand{\norm}[1]{\left\lVert#1\right\rVert}
\newcommand{\orcid}[1]{\href{https://orcid.org/#1}{\includegraphics[width=10pt]{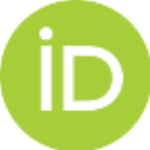}}}
\begin{document}
\title{Pickalo: Leveraging 6D Pose Estimation \\ for Low-Cost Industrial Bin Picking}
\author{Alessandro Tarsi$^\text{\orcid{0009-0001-7363-8289}}$*, Matteo Mastrogiuseppe$^\text{\orcid{0009-0000-9186-3693}}$*, Saverio Taliani$^\text{\orcid{0009-0003-9460-8913}}$*, Simone Cortinovis$^\text{\orcid{0009-0007-3599-2096}}$*, Ugo Pattacini$^\text{\orcid{0000-0001-8754-1632}}$
\thanks{*Alessandro Tarsi, Matteo Mastrogiuseppe, Saverio Taliani, Simone Cortinovis equally contributed to this work. This work was supported by Camozzi Automation SpA. Alessandro Tarsi, Matteo Mastrogiuseppe, Simone Cortinovis were with the MESH Facility (formerly iCub Tech), Istituto Italiano di Tecnologia, Genova 16163, Italy. Saverio Taliani was with the AMI laboratory, Istituto Italiano di Tecnologia, Genova 16163, Italy. Alessandro Tarsi is now with Institut des Systèmes Intelligents et de Robotique (ISIR), Paris 75005, France (email: tarsi@isir.upmc.fr). Matteo Mastrogiuseppe, Saverio Taliani, Simone Cortinovis are now with Generative Bionics, Genova 16152, Italy (email: \{matteo.mastrogiuseppe, saverio.taliani, simone.cortinovis\}@gbionics.ai). Ugo Pattacini is with the MESH Facility (formerly iCub Tech), Istituto Italiano di Tecnologia, Genova 16163, Italy (email: ugo.pattacini@iit.it).}
}

\makeatletter
\let\old@ps@headings\ps@headings
\let\old@ps@IEEEtitlepagestyle\ps@IEEEtitlepagestyle
\def\confheader#1{%
  \def\ps@IEEEtitlepagestyle{%
    \old@ps@IEEEtitlepagestyle%
    \def\@oddhead{\strut\hfill#1\hfill\strut}%
    \def\@evenhead{\strut\hfill#1\hfill\strut}%
  \def\@evenfoot{}
  }%
  \ps@headings%
}
\makeatother

\maketitle

\begin{abstract}
Bin picking in industrial environments remains challenging due to the high cost of industrial 3D sensors and the typical requirement of structured-light projectors. We present Pickalo, a modular pipeline designed to reduce the setup complexity of industrial bin-picking by leveraging 6D pose estimation and deep stereo matching. To maximize scene coverage, a wrist-mounted RGB-D camera actively explores the bin from multiple viewpoints. At each viewpoint, raw stereo streams are processed using BridgeDepth to obtain refined depth maps suitable for accurate collision reasoning. Object instances are segmented with a Mask-RCNN model trained on photorealistic synthetic data and localized using the zero-shot SAM-6D pose estimator. A pose buffer module fuses the observations over time, handling object symmetries and reducing pose noise. Offline, we generate and curate large sets of antipodal grasp candidates per object, while online, a utility-based ranking and fast collision checking are queried for the grasp planning. Deployed on a UR5e with a parallel-jaw gripper and an Intel RealSense D435i, Pickalo achieves up to 600 mean picks per hour with 96--99\% grasp success over 30 min runs on densely filled euroboxes. Ablation studies demonstrate the benefits of enhanced depth estimation and of the pose buffer for long-term stability and throughput in realistic industrial conditions. Videos are available at \href{https://mesh-iit.github.io/project-jl2-camozzi/}{https://mesh-iit.github.io/project-jl2-camozzi/}
\end{abstract}


\section{Introduction}

\begin{figure}[t]
\setlength{\belowcaptionskip}{-10pt}
    \centering
    \includegraphics[width=0.8\linewidth]{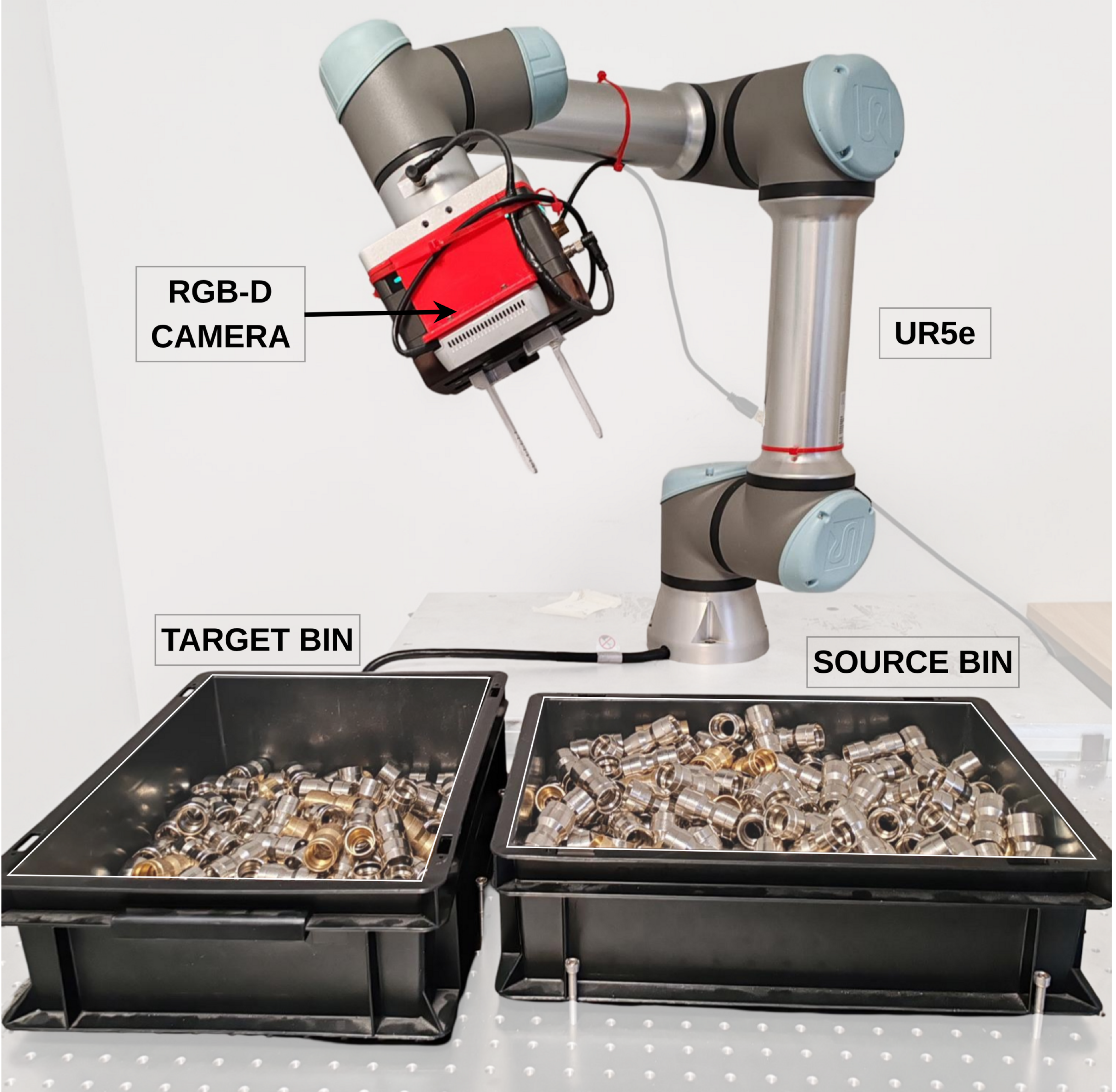}
    \caption{\small The experimental setup consists of a UR5e manipulator with a consumer-grade camera attached to the wrist. The~bin is a standard eurobox heavily filled with small metallic~objects.}
    \label{fig:setup}
\end{figure}

\IEEEPARstart{D}{espite} decades of research, bin picking remains a central challenge in industrial automation. The~task requires a robot to detect, localize, and~extract objects from a cluttered container under varying object poses. Conventionally, bin-picking systems adopt classical vision techniques such as 2D--3D feature matching, template matching, and~vote-based pose estimation~\cite{Rodrigues2012, Drost2010, VincentLepetit, CAD_based_recognition, Cad_based_pose}. These approaches are either not robust enough or assume access to high-quality depth data, typically obtained from industrial-grade stereo cameras. Such sensors produce dense point clouds with rich geometric detail, enabling registration algorithms to reliably align object models with observations, even under partial occlusion. However, these industrial-grade sensors require a complex and expensive setup, posing a barrier to user~adoption. Transformer-based architectures paved the way for bin-picking applications with lower-fidelity sensors and less constrained environments~\cite{Sun2024, Region-Aware, Peiyuan}. The~latest foundation models for pose estimation~\cite{Caraffa2025, wen2024foundationposeunified6dpose, Lin2023SAM6D} have demonstrated strong generalization to unseen object categories with minimal task-specific setup. On~the other hand, deep stereo matching models~\cite{wen2025foundstereo, Defomo, guan2025bridgedepth} aim to reconstruct accurate depth images, effectively narrowing down the gap between consumer-grade cameras and specialized high-end~hardware. As the performance of pose estimation algorithms is highly dependent on the depth quality~\cite{Probabilistic_Multi_View_Fusion, Robi}, in~this work, we focus on enhancing the quality of the depth input. Furthermore, we propose a pose buffer mechanism to increase the robustness of pose estimation and prove that consumer-grade cameras can be adopted in industrial bin picking without sacrificing~performance. The presented grasping pipeline follows a modular architecture, favoring available open-source packages when possible. Accuracy and timing constraints imposed by an industrial bin-picking application are explicitly considered. We conduct extensive and realistic evaluations across three different classes of objects. Moreover, we evaluate the robustness of our solution under industrial operating conditions, adopting standard eurobox bins loaded at their nominal capacity. Figure~\ref{fig:setup} illustrates the experimental setup adopted for the experiments.  Key contributions of this paper are:


\begin{itemize}
    \item \textbf{System integration:
    } We design a modular bin-picking system that integrates consumer-grade RGB-D sensing with existing depth-enhancement and zero-shot 6D pose-estimation foundation models, together with perception-aware grasp planning. 
    \item \textbf{{New mechanism:}} We propose the pose buffer, a~temporal multi-view pose-fusion mechanism that associates observations across viewpoints, explicitly accounts for object symmetries, and~filters unconfirmed estimates. It improves pose consistency and accuracy under occlusion. The~pose buffer internally builds the estimated state of the bin, retaining information regarding the multi-view consistency of poses and the presence of objects not visible from the current point of view.
    \item \textbf{{Experimental validation:}} We provide ablation studies of depth enhancement and pose buffer and~validate the full system in highly cluttered industrial bin-picking experiments over extended operating cycles, achieving 96--99$\%$ picking success across the evaluated objects.
\end{itemize}



\section{Related Works}
\label{sec:related_works}
\noindent This section reviews state-of-the-art methods for robotic bin picking. We analyze their strengths and limitations with respect to industrial constraints such as clutter, sensor cost, and~deployment~complexity. 

\subsection{Grasp Detection Algorithms}

\noindent Grasp detection approaches aim to directly predict feasible grasp poses for a robot gripper from sensor input. As~they bypass explicit object modeling and can be trained to map sensor data directly to grasp poses, end-to-end grasp detection methods are particularly attractive in cluttered, variable bin-picking settings.  Bui~et~al.~\cite{bui2024deep6d} present a network that generates six-DoF antipodal grasps from a point cloud of a cluttered bin, using a convolutional neural network (CNN) to identify and rank possible grasp poses. Sun~et~al.~\cite{Sun2024} propose a grasp detection framework that performs industrial bin picking using a low-cost sensor. Fang~et~al.~\cite{fang2023anygrasp} take a different approach with AnyGrasp, which fuses spatial and temporal information to improve grasp robustness across sequential frames. Despite the promising results, all these solutions were tested in simplified cases that do not resemble industrial scenarios. In~addition, the~goal of these methods is usually to find any stable grasp, which is insufficient when the downstream task demands a specific pickup orientation for quality control or precise placement. This motivates our decision to choose an alternative strategy based on the full six-DoF pose estimation of the~objects.

\subsection{6D Object Pose~Estimation}

\noindent Explicitly estimating object poses provides a clearer and more comprehensive description of the bin's state, which can be leveraged to plan grasps more effectively. In~particular, explicit pose estimation allows pose refinement over time and outlier detection based on time~consistency. Convolution Neural Networks have proven to be effective for 6D pose estimation~\cite{xiang2018posecnn, tekin2018realtime, Yann2022Megapose}. Li~et~al.~\cite{li2022s2rpick} integrated the CNN-based pose estimator proposed in~\cite{wang2019densefusion} to grasp texture-less metallic parts relying solely on simulation data. Their sim-to-real pipeline (S2R-Pick) automatically renders photorealistic scenes, producing a vast labeled dataset for training. However, the~dependence on large training datasets hinders the usability and scalability of this solution. Foundation models~\cite{Caraffa2025, wen2024foundationposeunified6dpose, Lin2023SAM6D}, on~the other hand, can perform zero-shot pose estimation and do not require a training step when the target object is changed, facilitating their adoption. Our grasping pipeline is based on SAM-6D~\cite{Lin2023SAM6D}, an~open-source, versatile, and~lightweight foundation model to estimate poses of novel~objects.

\subsection{Multi-View Consistency and Depth~Quality}
\noindent Industrial objects are often metallic and textureless, posing a problem for consumer-grade depth cameras.  For this reason, structured-light projectors and expensive optical systems are typically employed to mitigate depth degradation caused by light reflections. Recent works~\cite{Probabilistic_Multi_View_Fusion, fu2024lowcost} have proposed adopting multi-view strategies to improve the quality of the perceived depth. However, in~this work, we prove emerging foundation models for deep stereo matching are simpler and more effective. BridgeDepth by Guan~et~al.~\cite{guan2025bridgedepth} bridges monocular and stereo depth estimation, aligning context information given by mono-depth estimation with classical stereo disparity to produce dense and accurate depth maps. During~our experiments, we observed that adopting BridgeDepth reduced depth noise and reflection-related problems, increasing the number of detected objects and reducing the collisions caused by the presence of unseen objects. Furthermore, in~contrast to multi-view acquisition, deep stereo matching does not increase the complexity of the acquisition scheme in terms of robot motion and precise camera synchronization. To~keep pose estimation consistent across time, we employ a pose buffer module (Section~\ref{subsec:pose_memory}), that handles object symmetries as done in~\cite{labbe2020cosypose}. In~\cite{Huang_2025_ICCV}, a~multi-view approach was adopted before each grasp attempt to mitigate the risk of object miss-detection. In~our solution, instead, we enforce multi-view consistency across multiple grasping iterations while the system operates. The~presented approach allows Pickalo to reach a mean pick per hour (MPPH) of 600 objects, with~a success rate over $96\%$ in dense clutter using only a consumer-grade~camera.

\section{Methodology}
\begin{figure*}[htp]
    \centering
    \setlength{\belowcaptionskip}{-15pt}
    \includegraphics[width=1.05\textwidth]{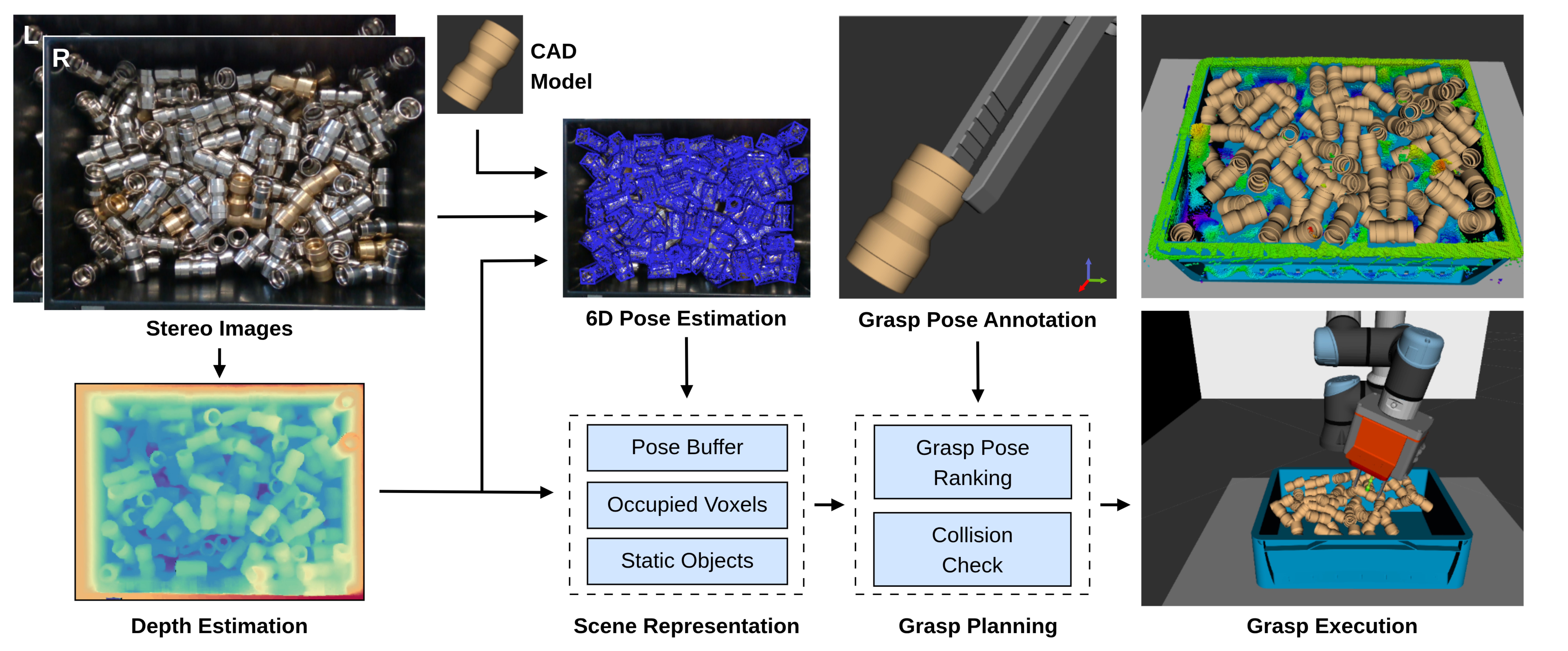}
    \caption{\small Overview of the presented pipeline. A~stereo-pair image is acquired and processed by the depth estimation block to obtain an enhanced depth reconstruction. The~resulting depth is aligned to the left RGB frame and provided to the 6D pose estimation model, together with the object CAD model. The~scene state is reconstructed by combining pose estimates across multiple views, occupied voxels, and~static objects. An~example of the scene representation is shown in the top-right corner. Given the scene state, the~grasp planning ranks and tests pre-computed grasp annotations to find a collision-free grasping~trajectory.}
    \label{fig:pipeline}
\end{figure*}

\noindent Our approach is illustrated in Figure~\ref{fig:pipeline}. The~pipeline presents a modular architecture that adopts state-of-the-art perception models and proposes simple but effective solutions for industrial bin picking. In~this section, we illustrate the key components of our architecture and the reasons behind their~success.

\subsection{Pose Camera selection}
\noindent In contrast to classical industrial scenarios relying on fixed cameras, we adopt an eye-in-hand camera mounted on the robot’s wrist. In~our experience, viewpoint variation helps to mitigate occlusion and ensure full bin scanning coverage. We employ a strategy that samples camera poses, at~optimal sensor distance, from~a hemisphere centered at the bin origin.
To ensure uniform coverage and viewpoint variability, reachable camera poses are ordered, maximizing the distance from previously selected~viewpoints. 

\subsection{Depth Map Acquisition}

\begin{figure}[h!]
  \centering
  {\includegraphics[width=0.24\textwidth]{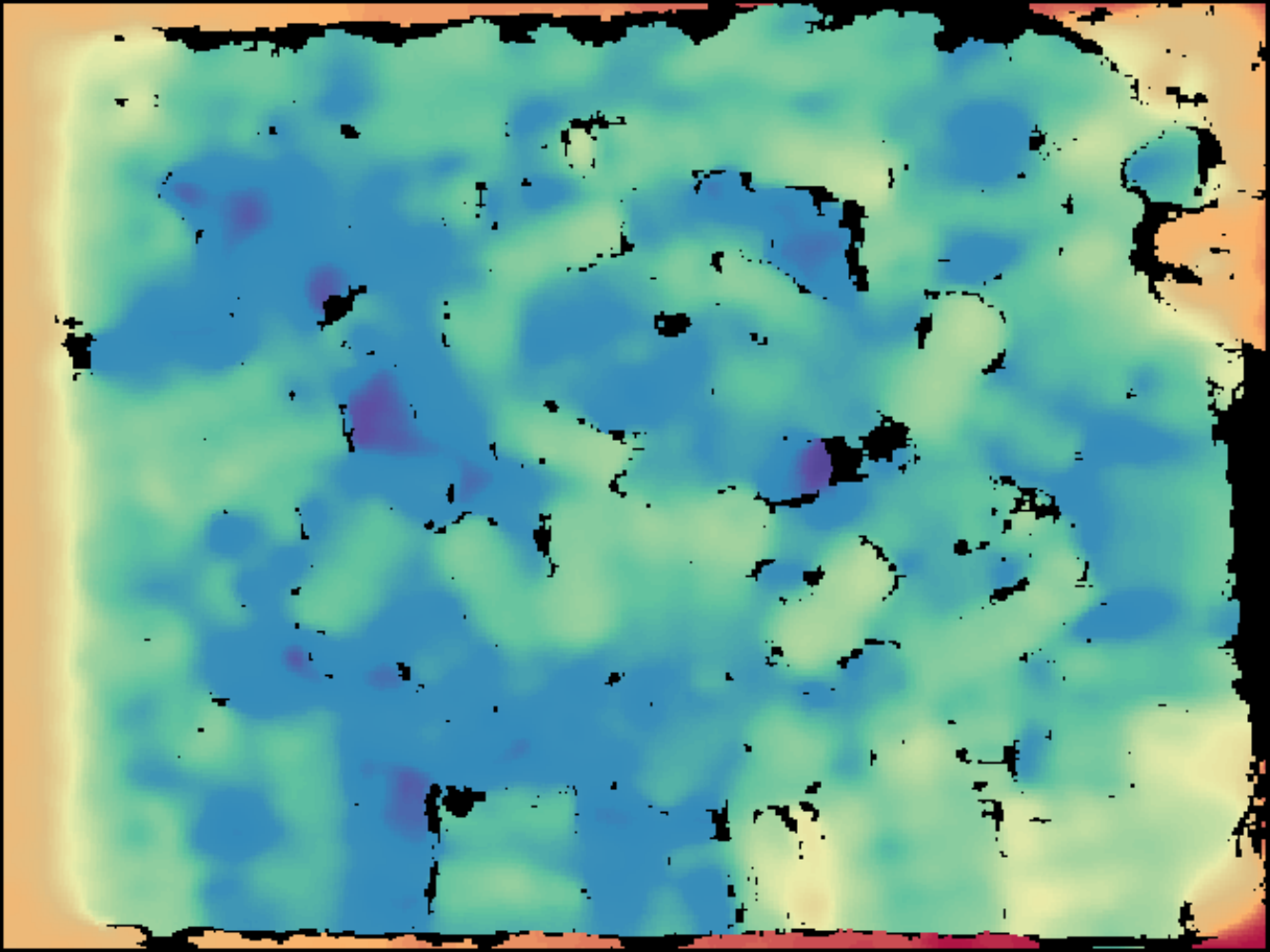}}
  \hfill
  {\includegraphics[width=0.24\textwidth]{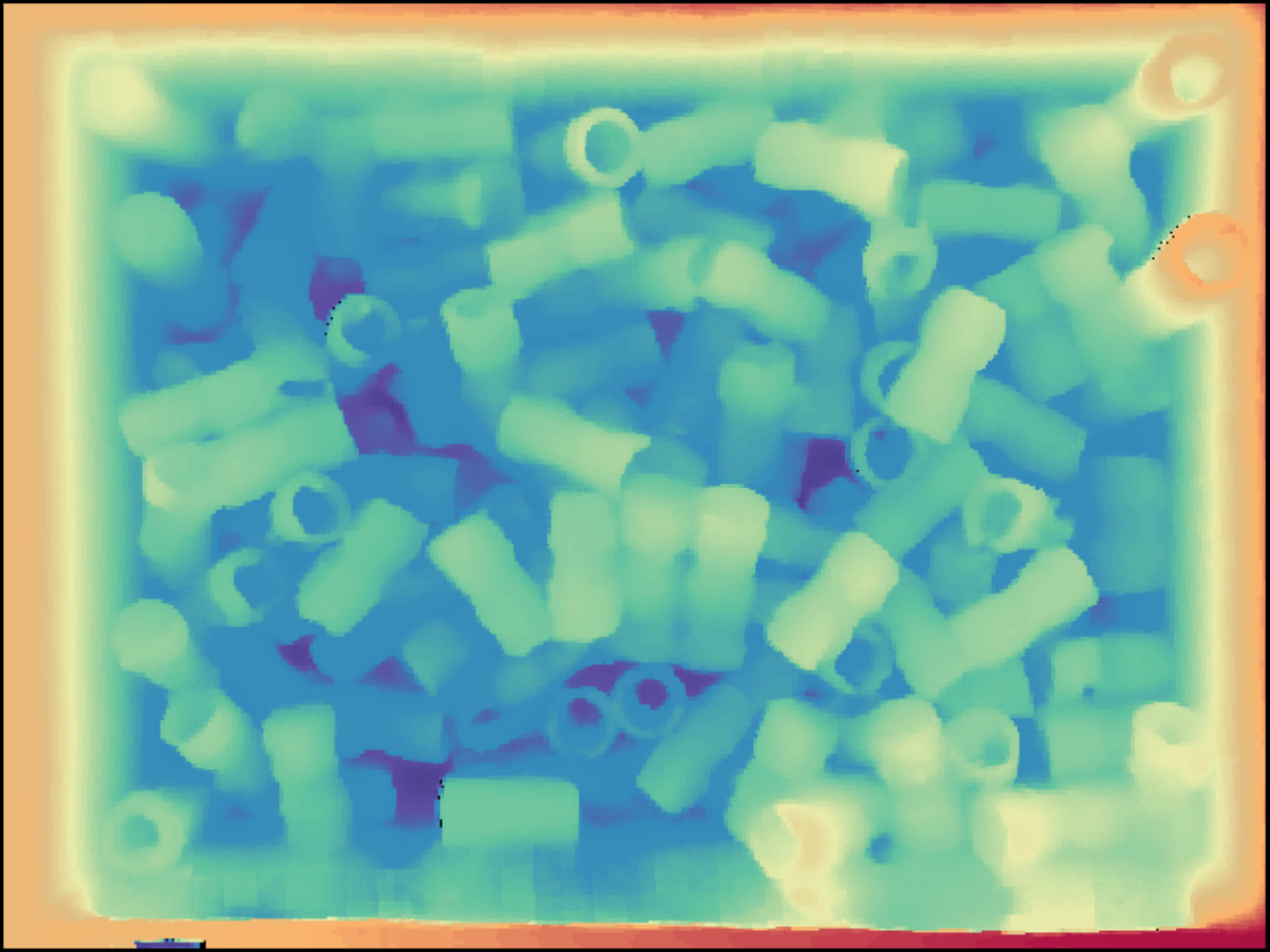}}
  \caption{\small RealSense on the left. BridgeDepth on the right.}
  \label{fig:depthcompar}
\end{figure}

\noindent Standard RGB-D sensors often produce inaccurate depth maps, especially in the presence of reflective metallic objects~\cite{fu2024lowcost}.
To address this problem, we adopt BridgeDepth, a~lightweight deep stereo matching framework that merges the advantages of stereo matching with the context understanding of monocular depth estimation~\cite{guan2025bridgedepth}. Our experiments demonstrate that BridgeDepth produces sufficiently accurate depth maps and is robust to variations in lighting conditions. A~qualitative depth comparison is shown in Figure~\ref{fig:depthcompar}. Moreover, being open source and with a striking accuracy-runtime trade-off, BridgeDepth is a perfect candidate for our grasping pipeline. To~the best of our knowledge, this work is the first one to test the effect of deep stereo matching on an industrial bin-picking~scenario. 

\noindent BridgeDepth outperforms alternatives, such as the multi-view acquisition proposed in~\cite{fu2024lowcost}, in~which a more accurate depth map was reconstructed from multiple depth images and a per-pixel voting system. In~our experiments, multi-view depth estimation proved to be slower, due to the extra robot motion, and~less accurate, as~it targeted reflection inaccuracies without improving general depth~quality.

\subsection{6D Object Pose Estimation}
\noindent Object pose estimation (OPE) is the core element of our robotic bin-picking applications. Our perception pipeline is based on SAM-6D~\cite{Lin2023SAM6D}, an~open-source approach for 6D pose estimation of unseen objects. Due to the time constraints imposed by the industrial requirements, the~original implementation of SAM6D was adapted as described~below.

\noindent \textbf{{Instance Segmentation.}} SAM-6D is based on Segment Anything Model~\cite{SAM}, a~zero-shot foundation model for instance segmentation that, despite its generalization capabilities, proves to be too slow for industrial purposes. As~runtime performance is key in industrial scenarios, we substituted the instance segmentation model with a custom Mask R-CNN backbone. This choice makes an upfront training step required for the segmentation backbone. To~facilitate usability, we created an automated pipeline that, starting from the CAD model, generates a training dataset using BlenderProc~\cite{BlenderProc} and trains the Mask R-CNN backbone~\cite{MASKRCNN}. Starting from each CAD model, our automated BlenderProc pipeline generates approximately 1000 synthetic images per object class. For~each scene, multiple instances are initialized at random positions and orientations above the bin and allowed to settle under gravity, producing physically plausible poses, dense stacking, and~occlusions. The~Mask R-CNN  model with a ResNet50 backbone is fine-tuned for 20 epochs using SGD as optimizer, a~learning rate of 0.00125, a~batch size of 8 and an input resolution of (640,~480).

\noindent \textbf{{Pose Estimation.}} We adopted the same pose estimation module presented in SAM-6D, as~our experiments proved this method to be reliable, fast, and~accurate. However, to~improve the effectiveness of the system under severe clutter, we introduced a simpler but effective pose rejection~filter.

\noindent \textbf{{Pose Rejection Filter.}} The pose rejection filter verifies whether an estimated object pose is physically consistent with the observed depth data. 
Let $BB_c = (u_c,v_c)$ be the center of a bounding box in the image plane and~$z_{mean}$ the mean depth value of pixels within the corresponding segmentation mask. We project the bounding box center into 3D space using the camera matrix $K$:
\begin{equation}
    ^{c}t_{BB} = z_{mean} K^{-1} \left[u_c, v_c, 1\right]^T
    \label{project_BB_C}
\end{equation}
$^ct_{BB}$ expresses the mean distance of the visible part of the object perceived by the depth image.
Let $^{c}t_{obj}$ be the translation vector of the estimated centroid of the object expressed in the camera frame. A~pose candidate is rejected if:
\begin{equation}
\norm{ ^{c}t_{BB} } < \norm{ ^{c} t_{obj} }.
\end{equation}
where $\norm{\cdot} $ represents the Euclidean norm. The~pose rejection filter effectively removes estimated poses for which the segmented point cloud is matched with the rear surface of the object, leading the estimated pose to be closer to the camera than the observed depth map. During~experiments, this error was observed any time an object buried in the bin was surrounded by neighbor objects, causing the model to mistakenly consider the lateral faces of the neighbors as the lateral faces of the target~object.

\subsection{Pose~Buffer}
\label{subsec:pose_memory}
\noindent To create a consistent scene representation relying on the imperfect pose estimations, we introduce the pose buffer. The~pose buffer is a temporal buffer that associates pose estimates of the same object across multiple views over different time instances. Let ${^w H_{i}^k} \in SE(3)$ be the homogeneous transformation describing the pose of object $i$ at iteration $k$ in the world frame. For~simplicity, we denote $^w H_{i}^k$= $p_i^k = (R_i^k,t_i^k)$.

\noindent \textbf{Pose Averaging}. The~pose buffer aggregates a history of estimates $\{p_i^0, \dots, p_i^{k}\}$ and computes a more robust pose estimate $\widehat{p}_i$ using Euclidean and quaternion averaging~\cite{quaternion_averaging} for translation and rotations, respectively. 

\noindent \textbf{Data Aggregation}. An~incoming pose $p^{k} = (R^k,t^k)$ is associated with an existing object $i$ in memory if:
\begin{equation}
\begin{aligned} \label{eq:PM_condition}
\cos^{-1} \left((\operatorname{Tr}(R^k \widehat{R}_i^T) - 1)/2\right)    &< \theta_{thresh}  \\ 
\norm{t^k - \widehat{t}_i}_2 &< \delta_{thresh}
\end{aligned}
\end{equation}
where $\widehat{p}_i = (\widehat{R}_i,\widehat{t}_i)$ is the estimated pose of object $i$ at the current state, $Tr(\cdot)$ is the trace operator, and~$\theta_{thresh}$ and $\delta_{thresh}$ represent the maximum angular and Euclidean distances, respectively. In~our experiments, we used 10 degrees and $0.0075$ meters as values for $\theta_{thresh}$ and  $\delta_{thresh}$, respectively.

\noindent \textbf{Object Symmetries}. The~handling of the object symmetries draws inspiration from the work done in~\cite{labbe2020cosypose,pitteri2019object}. We define $\mathcal{R}(l, p)$ as the visual appearance of object $l$ rendered in pose $p$. The~symmetry group $S(l)$ is defined as the set of transformations that leave the visual appearance invariant:
\begin{equation}
    S(l) = \{ S \in SE(3) \mid \forall p, \mathcal{R}(l, p) = \mathcal{R}(l, S \cdot p) \}
\end{equation}

\noindent $S(l)$ is automatically computed offline starting from the object’s model. In~particular, we rotate the vertices composing the object’s mesh around its axes and verify whether the resulting cloud of vertices is identical to the initial~one.

\noindent To find out if a new pose estimate $p^{k}$ matches one of the $m$ previously detected object poses present in the memory, the~system checks if:
\begin{equation}
\begin{aligned} 
\exists i&\in\{0,\ldots,m-1\}, S\in S(l) \text{ such that } \\ 
&\mathcal{R}(l,\widehat{p}_i)=\mathcal{R}(l,S\cdot p^k) 
\end{aligned}    
\end{equation}

\noindent If a match is found, the~new pose $p^{k}$ is added to the list of previous estimates $\{p_i^0, \dots, p_i^{k-1}\}$ of object $i$, and~$\widehat{p}_i$ is~updated. 

\noindent \textbf{Pose validation}. The~pose buffer exposes only the pose estimates linked to objects observed in the latest iteration, and~for which there are at least two sample observations. In~addition, pose estimates are subject to invalidation if new observations do not confirm the presence of the object. These mechanisms ensure that the poses used in downstream tasks are reliable and up-to-date, e.g.,~they have not been invalidated during the latest object~extraction.

\subsection{Scene~Representation}
\noindent The system combines the information coming from the pose buffer and the depth estimation to build an accurate bin state representation. This representation is composed~of:

\begin{enumerate}
    \item \textbf{Target Objects:} The mesh of the objects of interest at their estimated locations according to the pose buffer.
    \item \textbf{Static Objects:} Whose poses are determined through an initial calibration, such as the table and the bin.
    \item \textbf{Occupied Voxels:} Regions of space detected as occupied by the estimated depth but not associated with the former two categories.
\end{enumerate}

\noindent The scene representation is used by the grasp planning to access the validated pose estimates, to~access the collision information, and~to generate valid grasping~trajectories. 
 
\subsection{Grasp~Planning}

\noindent From each estimated object pose, we test possible offline-generated grasp candidates, which are then converted into collision-free robot trajectories for object~extraction. 

\noindent Grasp planning is divided into three stages: offline generation of grasp candidates, online ranking, and~collision~check.

\noindent \textbf{Generation of Grasp Candidates}. Grasp candidates are generated offline for each object category to create a database of feasible end-effector poses. The~generation process draws inspiration from~\cite{kleeberger2021automaticgraspposegeneration} and~follows a geometry-driven approach:

\begin{enumerate}
    \item \textbf{Antipodal Sampling:} We specify a target number $N$ of grasp candidates. The~algorithm samples points on the object's mesh surface and performs ray-tracing along the direction of the surface normals. A~pair of points is considered a candidate for grasping if their normals are antipodal within a specified angular tolerance, ensuring a stable force-closure~grasp.
    
    \item \textbf{Frame Definition:} For every identified antipodal pair, the~corresponding gripper pose is computed. The~closing axis is defined by the line connecting the contact points. The~approach vector is constrained to be orthogonal to the closing axis and oriented outward, away from the object's geometric~center.
    
    \item \textbf{Mesh Collision Check:} Finally, the~algorithm validates the candidate pose against the specific geometry of the gripper. We perform a collision check between the CAD model of the gripper fingers and the object mesh. To~ensure a collision-free approach, candidates resulting in an intersection between the gripper and the object are discarded.
\end{enumerate}

\noindent The resulting dataset of generated grasp candidates can be easily inspected and manually refined to accommodate specific industrial~needs.

\noindent \textbf{Grasp Ranking Strategy}. During~the online phase, we instantiate candidate grasps for every detected object instance. Since checking inverse kinematics (IK) and collisions for thousands of candidates is computationally expensive, we rank the candidates to prioritize those most likely to succeed.
We define a utility score $S(g)$ for a grasp $g$, computed as a weighted sum of four metrics:
\begin{equation}
    S(g) = w_1 S_{align} + w_2 S_{yaw} + w_3 S_{conf} + w_4 S_{height}
\end{equation}
where $w_1, w_2, w_3, w_4$  are weights in $[0,1]$, and their sum is equal to~1.

\begin{figure}[t]
    \includegraphics[width=\linewidth]
{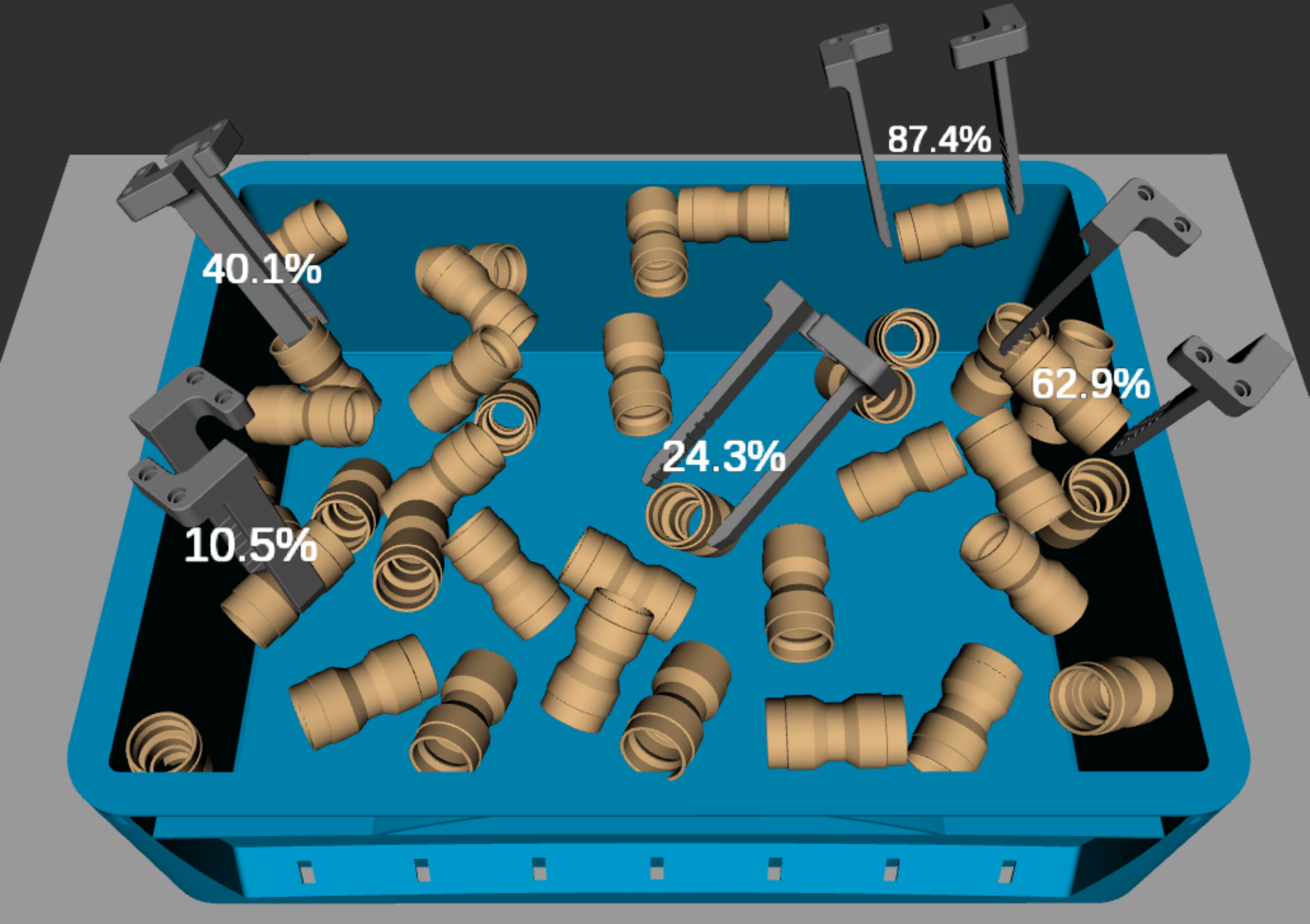}
    \caption{Example of various grasping poses with the corresponding utility scores $S(g)$ given in percentage. The~grasping poses are sorted in descending~order.}
    \label{fig:ranking}
\end{figure}

\begin{itemize}
    \item \textbf{Vertical Alignment ($S_{align}$):} We prioritize grasps where the approach vector aligns with gravity. Let $z_{ee}$ be the gripper's approach vector expressed in the world frame and~$z_g$ the gravity unit vector.
\begin{equation}
        S_{align} = \dfrac{ 1 + z_{ee} \cdot z_g}{2}
    \end{equation}
    This term is maximized when the gripper approaches the object vertically~downwards.

    \item \textbf{Yaw Deviation ($S_{yaw}$):} We penalize the rotation of the gripper around the vertical axis to minimize the motion of the wrist joint. This is computed similarly to $S_{align}$ but~compares the gripper's lateral axis against the bin's principal~axis.
    
    \item \textbf{Pose Confidence ($S_{conf}$):} A normalized score derived from the pose buffer, prioritizing objects detected with higher~certainty.
    
    \item \textbf{Stacking Height ($S_{height}$):} A normalized height value, prioritizing objects on top of the pile to minimize possible collisions.
\end{itemize}

\noindent \textbf{Collision Check}.~Given the ranked list of grasping poses, we want to generate a collision-free trajectory to extract the target object. To~ensure safety while maintaining a high success rate, the~collision check module implements a two-stage hierarchical validation process:

\begin{figure}[t]
    \includegraphics[width=\linewidth]
{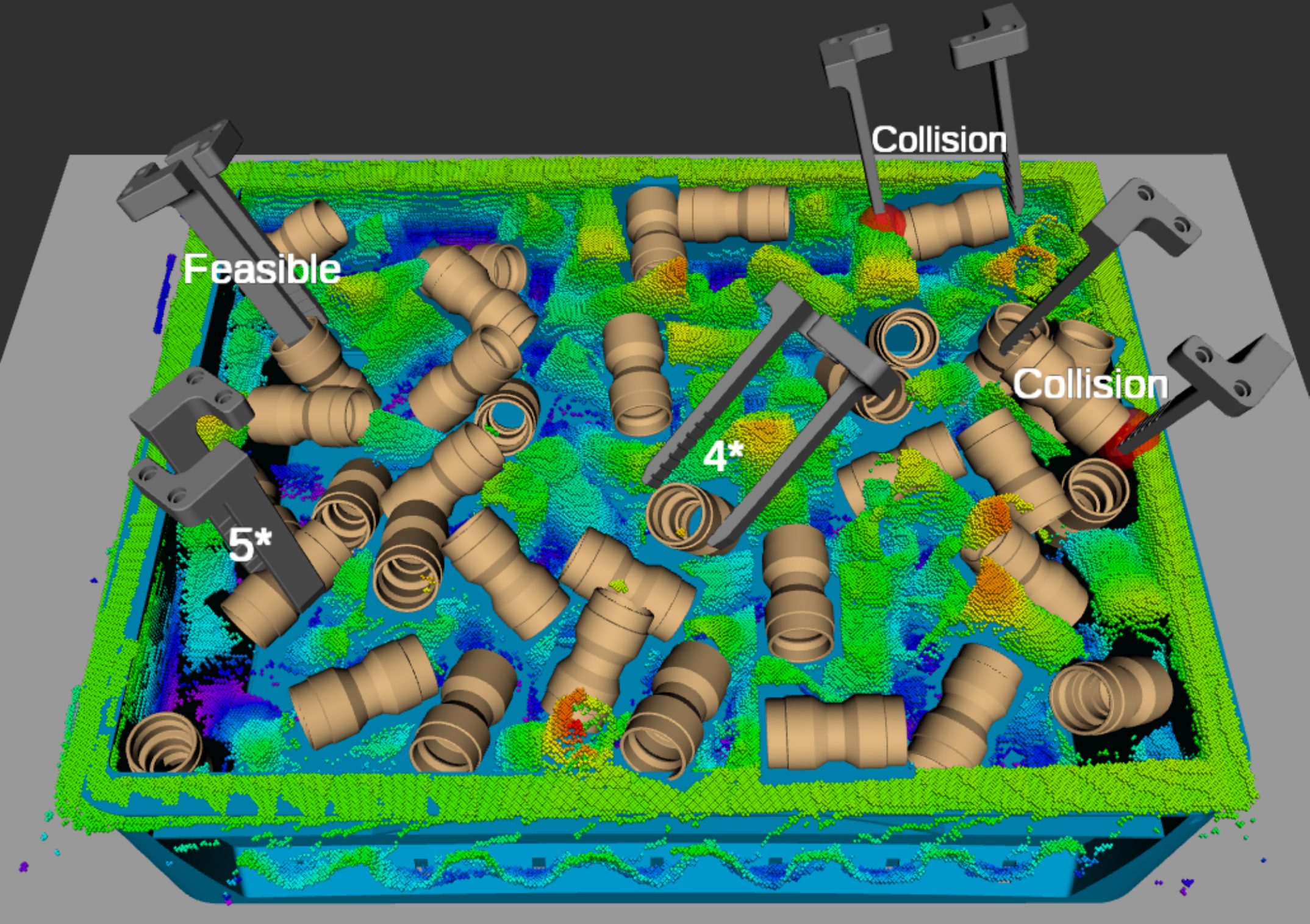}
    \caption{The sorted grasp poses are checked following the ranking order until a feasible grasp pose is found. Grasp poses leading to a collision are~discarded.}
  \label{fig:pruning}
\end{figure}

\begin{enumerate}
    \item \textbf{Static Pose Validation:} 
    The system first verifies if the robot can reach the target grasp configuration  without any collisions. In~this stage, all collision constraints are active. The~robot must not collide with the bin, the~objects, or~the occupied voxels. If~this check fails, the~grasp candidate is immediately~discarded.

    \item \textbf{Full Trajectory Validation:} 
    For candidates passing the static check, a~motion planner generates a path to reach and extract the target object. During~this phase, strict collision checking is maintained against the static environment (bin and table). However, collision constraints with occupied voxels are relaxed. 
    This modification allows the gripper to slightly touch neighboring objects while reaching the target grasp pose. Given the nature of the metallic objects we are dealing with, we deem occasional incidental contacts acceptable.
\end{enumerate}

\noindent Experimental data demonstrate that if a valid solution exists, it is found within the top $18\%$ of the ranked grasps in $99\%$ of the cases. Based on this observation, we evaluate only the top $18\%$ of the ranked list. If~no solution is found within this subset, the~system marks the current scene as unfeasible for immediate grasping and triggers a re-acquisition routine from a new viewpoint. If~a feasible trajectory is found, the~search is terminated, and~the joint trajectory is sent to the robot controller, which executes it in open~loop.

\noindent During execution, failures are monitored through the position and velocity of the gripper’s fingers. If~slippage is detected, the~robot retracts, the~bin state is updated to reflect the failed attempt, and~the pipeline restarts with a new~acquisition.

\noindent Figures~\ref{fig:ranking} and \ref{fig:pruning} present, respectively, an example of the value of the utility score $S(g)$ for a selected subset of grasps and~the subsequent collision check that verifies trajectory validity from the highest ranked grasp to the lowest until a successful grasp is correctly~found.

\subsection{Masked~Time}
\begin{figure}[htp]
\setlength{\belowcaptionskip}{-15pt}
    \centering \adjincludegraphics[width=0.5\textwidth]{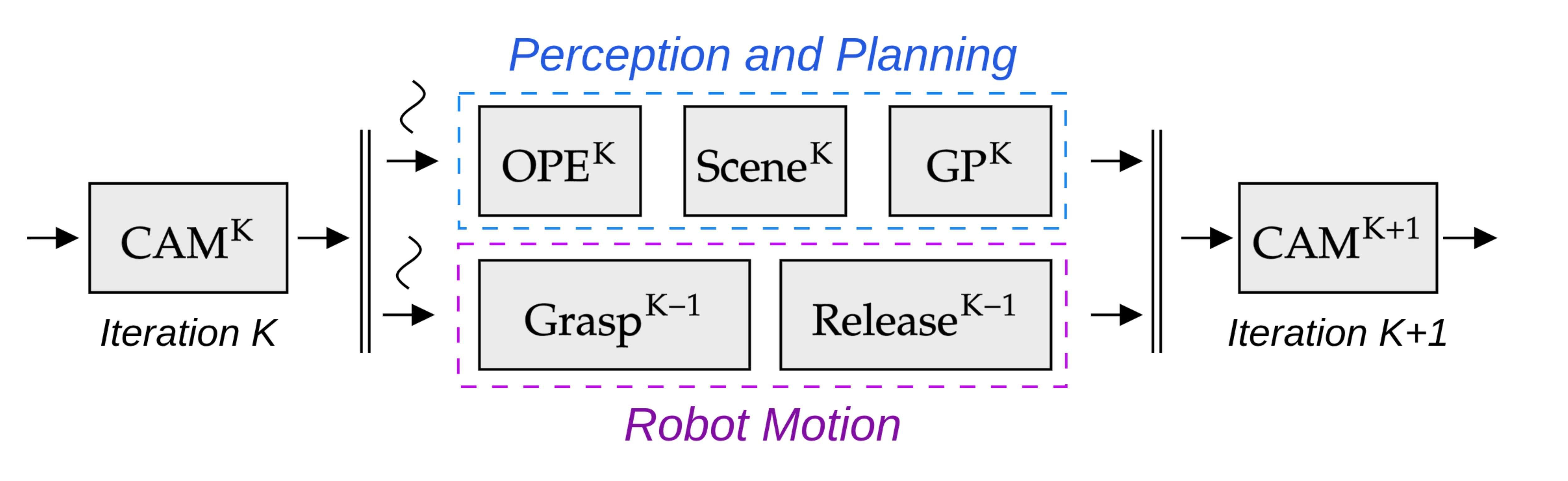}
    \caption{\small The grasping pipeline is parallelized to elaborate the perception and planning blocks for the current iteration, while the robot executes the grasping and releasing trajectories computed at the previous~iteration.}
    
    \label{fig:masked_time}
\end{figure}

\noindent To reduce idle time in our system, we distribute tasks across multiple threads. The robot motion is parallelized to the perception and planning phase, as~illustrated in Figure~\ref{fig:masked_time}. The~camera acquisition determines the beginning of a new iteration $k$. The~incoming new image is passed to the object pose estimation block, while the robot executes the grasping trajectory computed at the previous iteration step. During~the robot motion, once the estimated poses become available, the~scene representation is updated, and~the grasp planner is triggered. At~the end of the release phase, the~robot acquires a new image, and~a new cycle~begins.

\section{Experiments}

\subsection{Experimental Setup}

\noindent The experimental testbed consisted of the hardware components listed below. While specific models are detailed for reproducibility, the~pipeline's modular design enables the substitution of alternative components. Interestingly, the~system has already been tested on a platform with reduced computational power, which caused a minor impact on the cycle time due to masked time that leverages the robot motion to compensate for computational~delays.

\begin{itemize}
    \item \textbf{Manipulator:} Universal Robots UR5e (6-DoF).
    \item \textbf{Camera:} Intel RealSense D435i depth camera.
    \item \textbf{Computing:} Intel Core i9 (14th Gen), 128 GB RAM, and~RTX 4090 GPU.
\end{itemize} 

\begin{figure}[htbp]
\setlength{\belowcaptionskip}{-15pt}
  \centering
  \noindent\makebox[\columnwidth][c]{
    \begin{minipage}[t]{0.33333\columnwidth}
      \centering
      \includegraphics[width=\linewidth]{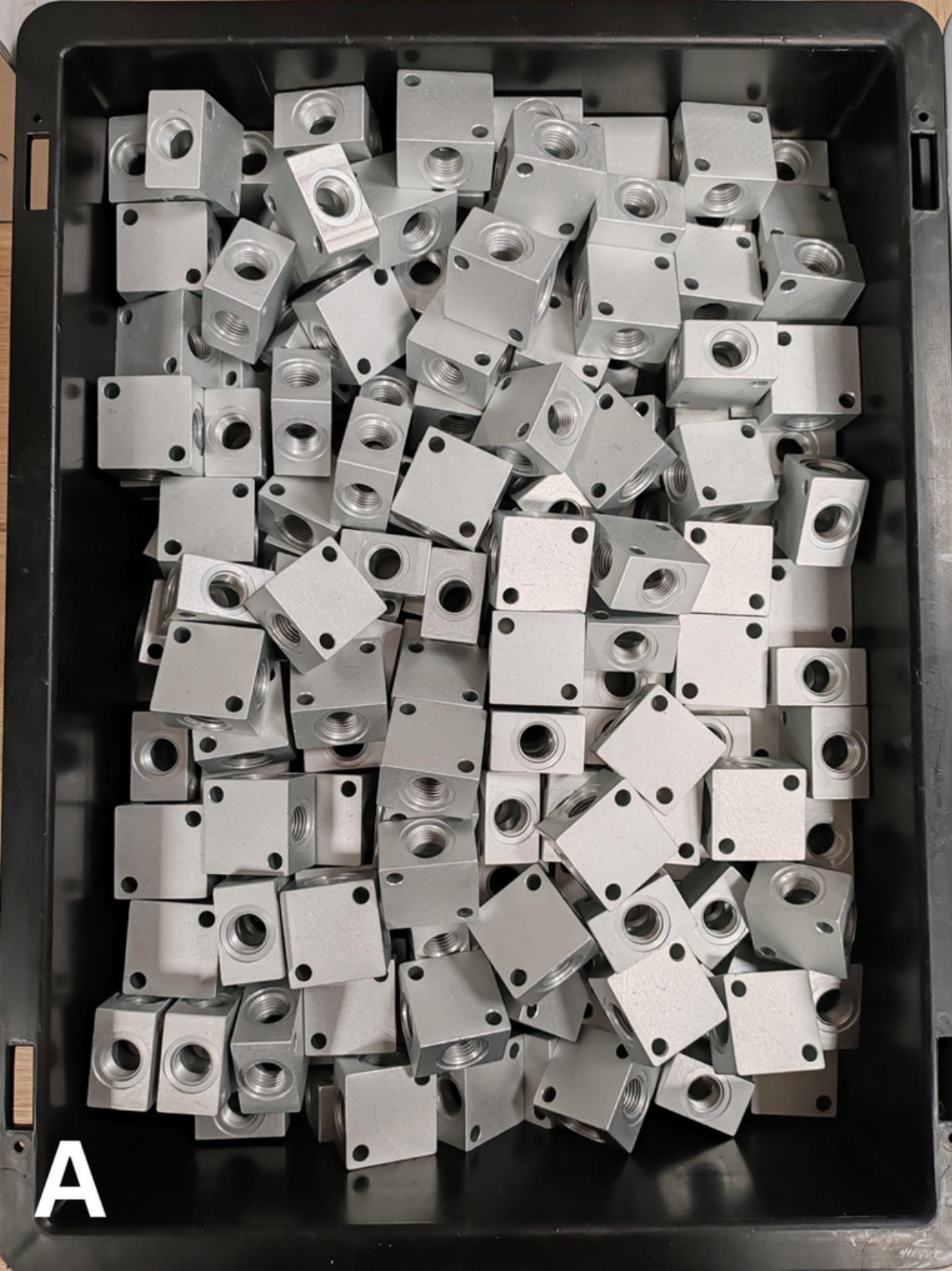}
      \label{img:cuboidi}
    \end{minipage}\hfill%
    \begin{minipage}[t]{0.33333\columnwidth}
      \centering
      \includegraphics[width=\linewidth]{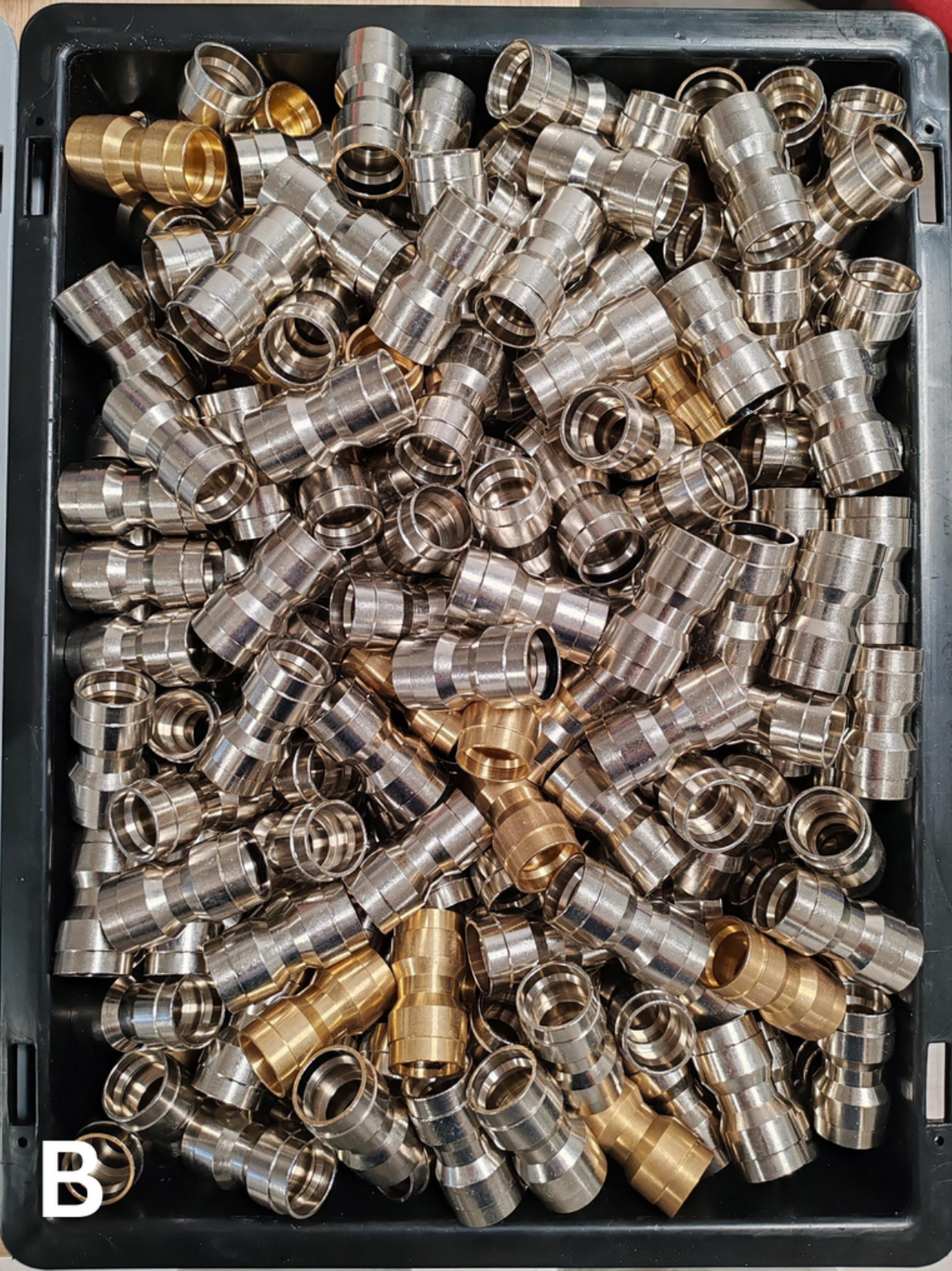}
      \label{img:ottonati}
      \end{minipage}
    \begin{minipage}[t]{0.33333\columnwidth}
      \centering
      \includegraphics[width=\linewidth]{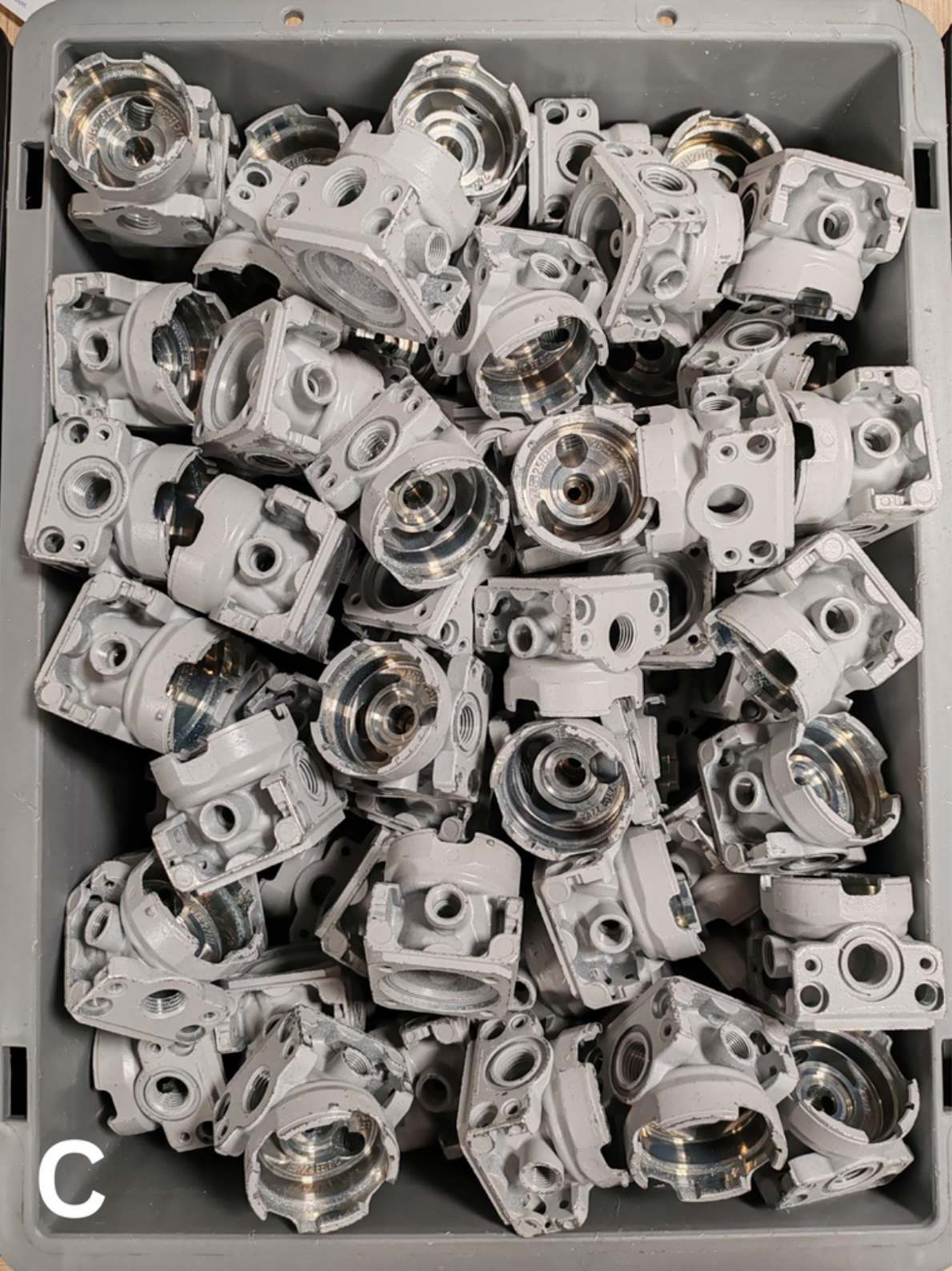}
      \label{img:frl}
    \end{minipage}%
  }
    \caption{\small Test objects used in the evaluation: (A) Square, (B) Cylindrical, and (C) Complex geometry. }
    \label{fig:objects}
\end{figure}

\noindent The system validation mirrored the actual complexity of industrial scenarios. Our setup was tested with no constraints on the bin state or ambient illumination, and~against arbitrarily full containers. This approach addresses a gap in the literature, where methods are often tested on sparse scenes, with~fixed lighting, or~without standard bins, effectively bypassing the complexities of object occlusion and environmental variability. As~illustrated in Figure~\ref{fig:objects}, we used three sets of industrial components featuring highly reflective surfaces and complex~geometries.

\subsection{Grasp Pipeline Performance}

\noindent To quantitatively evaluate the system's operational efficiency and grasp reliability, we adopted the following set of performance metrics:

\begin{itemize}
    \item \textbf{Mean Picks per Hour (MPPH):} A standard metric to evaluate system throughput.
    \item \textbf{Success Rate (SR):} The percentage of successful grasp executions over the total grasping attempts. We considered a grasp to be successful if the object was safely grasped from the source bin and released in the target bin, regardless of its final pose.
    \item \textbf{Early Exit Rate (EER):} The percentage of iterations where execution halted because no valid, collision-free grasp candidate could be identified.
\end{itemize}

\noindent Table~\ref{tab:three-obj-comparison} summarizes the results over 1,000 pipeline iterations. The~system maintains high throughput (MPPH) and very high reliability ($\text{SR} > 96\,\%$) across all categories; however, object geometry influences~performance. 

\begin{table}[ht]
\centering
\caption{\small Performance metrics across object categories (1,000 iterations).}
\label{tab:three-obj-comparison}
\begin{tabular}{cccc}
\toprule
\textbf{Obj.} & \textbf{MPPH} & \textbf{SR (\%)} & \textbf{EER (\%)} \\ 
\midrule
A & 577.85 & 98.84 & 4.22 \\
B & 615.89 & 97.73 & 0.62 \\
C & 611.83 & 96.12 & 2.98 \\ 
\bottomrule
\end{tabular}
\end{table}

\noindent Object A (square) offers stable grasp poses for a parallel-jaw gripper, resulting in a high SR. However, its shape causes objects to stack tightly, occluding collision-free approach vectors and leading to a higher EER. In~contrast, Object B (cylindrical) resists stacking, resulting in a minimal EER, though~the curved surface offers slightly less grasp stability. Finally, Object C represents the most challenging case due to its weight and complex geometry; while it offers many potential grasp poses, the~limited contact surface area in certain grasping configurations results in a lower overall success~rate.

\subsection{Ablation~Study}
\noindent To validate the individual contributions of specific components within our pipeline, we conducted an ablation study focusing on two key aspects: the depth estimation method and the impact of the pose~buffer.

\noindent \textbf{Depth Estimation.}~We compared the performance of raw depth input from the Intel RealSense D435i against the BridgeDepth method using Object B over three duration intervals (5, 10, and~30 min). The~results are summarized in Table~\ref{tab:depth-ablation}.

\begin{table}[ht]
\centering
\caption{\small Ablation: RealSense Depth (RS) vs. BridgeDepth (BD). Best result for each time interval is shown in bold.}
\label{tab:depth-ablation}
\setlength{\tabcolsep}{3.5pt} 
\small 
\begin{tabular}{clcccc}
\toprule
\textbf{Time} & \textbf{Method} & \textbf{MPPH} & \textbf{SR (\%)} & \textbf{EER (\%)} \\ 
\midrule
\multirow{2}{*}{5 min}  
 & RS & \textbf{629.9} & 96.9 & 0.9 \\
 & BD & 615.9 & \textbf{97.7} & \textbf{0.6} \\ 
\multirow{2}{*}{10 min} 
 & RS & 605.4 & 92.7 & 1.6 \\
 & BD & \textbf{621.6} & \textbf{97.7} & \textbf{0.3} \\ 
\multirow{2}{*}{30 min} 
 & RS & 557.5 & 88.5 & 4.9 \\
 & BD & \textbf{598.3} & \textbf{96.3} & \textbf{2.4} \\ 
\bottomrule
\end{tabular}%
\end{table}

\noindent The RealSense baseline exhibits slightly higher throughput (MPPH) during the first 5 min, primarily due to the absence of model inference latency. However, it suffers from significant performance degradation over longer durations. By~the 30 min mark, the~RealSense success rate (SR) drops to 88.5\%, while the Early Exit Rate (EER) reaches 4.9\%.  This degradation is attributed to the lower quality of the raw depth, which contains holes and artifacts. These inconsistencies lead to missed object detections and erroneous collision checks (higher EER), as~well as inaccurate pose estimations (lower SR). In~contrast, BridgeDepth demonstrates superior long-term stability. At~30 min, it maintains an SR of 96.3\% compared to 88.5\% of the~baseline.

\begin{figure}[ht]
\setlength{\belowcaptionskip}{-5pt}
    \centering
\includegraphics[width=1.0\linewidth]{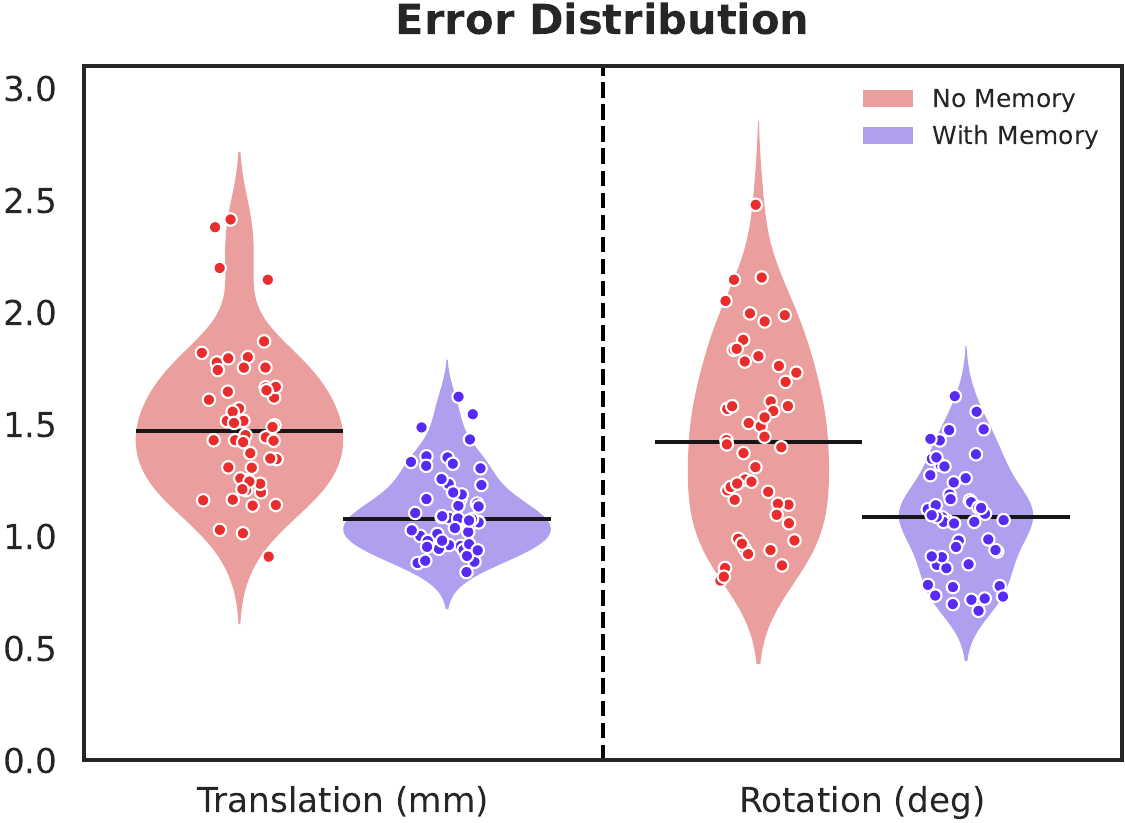}
    \caption{\small Error distribution of the object pose estimation, enabling and disabling the pose buffer module. The~translation and angular error represent the average distance between the estimated pose and the ground truth pose given by the XYZ-IBD~dataset.}
\label{fig:ablation_on_public_dataset}
\end{figure}

\noindent \textbf{Pose Buffer.} To measure the specific impact of the pose buffer module, we analyzed the system while enabling or disabling the pose buffer module; we named the two conditions ``With Memory'' and ``No Memory'', respectively. For the study we adopted the XYZ-IBD dataset~\cite{huang2025xyzibdhighprecisionbinpickingdataset}. The~XYZ-IBD dataset shares similarities with our setup, such as multiple identical textureless industrial parts, dense clutter, and~ambiguous object appearances. First, we evaluated the accuracy of the pose estimation using object class ``000010''. This object closely resembles the industrial objects used in our experiments. We estimated the poses of the object instances across 50 images without the temporal buffer, calculating the average translation and rotation errors relative to the ground truth. We then repeated the test, enabling the temporal buffer, and~fusing pose estimates from different viewpoints as described in Section~\ref{subsec:pose_memory}. 
The translation and rotation errors were computed using the same distance metrics introduced in the pose buffer association condition (Equation~(\ref{eq:PM_condition})). As~shown in Figure~\ref{fig:ablation_on_public_dataset}, the pose buffer reduces both the pose average error and the standard~deviation. Next, we compared the full pipeline performance using Object B over three time intervals (5, 10, and~30 min). The~results, shown in Table~\ref{tab:memory-ablation}, highlight the critical role of the pose buffer in optimizing grasp~execution.

\begin{table}[ht]
\centering
\caption{\small Ablation: No Memory (NM) vs. With Memory (WM). Best result for each time interval is shown in bold.}
\label{tab:memory-ablation}
\setlength{\tabcolsep}{3.5pt}
\small
\begin{tabular}{clcccc}
\toprule
\textbf{Time} & \textbf{Method} & \textbf{MPPH} & \textbf{SR (\%)} & \textbf{EER (\%)} \\ 
\midrule
\multirow{2}{*}{5 min}  
 & NM & 565.3 & 86.8 & \textbf{0.0} \\
 & WM & \textbf{615.9} & \textbf{97.7} & 0.6 \\ 
\multirow{2}{*}{10 min} 
 & NM & 557.9 & 84.9 & \textbf{0.0} \\
 & WM & \textbf{621.6} & \textbf{97.7} & 0.3 \\ 
\multirow{2}{*}{30 min} 
 & NM & 553.9 & 84.2 & \textbf{0.0} \\
 & WM & \textbf{598.3} & \textbf{96.3} & 2.4 \\ 
\bottomrule
\end{tabular}%
\end{table}

\noindent The inclusion of the temporal buffer module consistently boosts performance, increasing MPPH by 10--12\% and the success rate by 11--13 percentage points. Notably, the~``No Memory'' configuration exhibits a near-zero Early Exit Rate, which is a direct consequence of our scoring criteria. When the module is active, an~object is only considered graspable if it has been estimated at least twice, including in the most recent iteration. Conversely, without~memory, every pose estimation is immediately treated as a potential grasp. Consequently, while the ``No Memory'' system attempts to grasp more objects (yielding a lower Early Exit Rate), it encounters significantly more errors. This results in a lower success rate (84--87\%) compared to the system with memory ($>$96\%), which successfully filters out unreliable~attempts.

\subsection{Bin-Emptying~Performance}
\begin{figure}[t]
\setlength{\belowcaptionskip}{-5pt}
    \centering
    \includegraphics[width=1.0\linewidth]{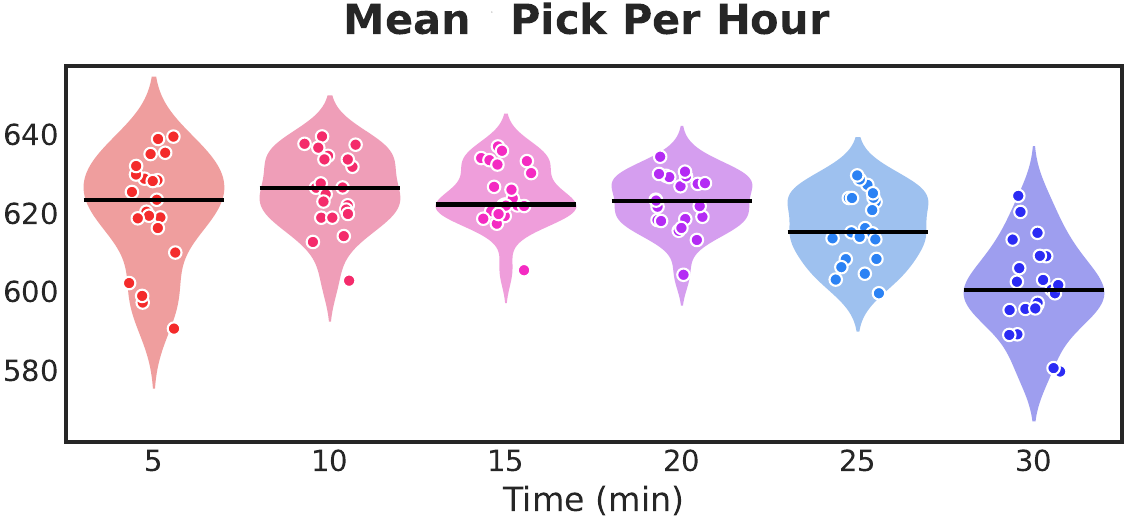}
    
    \vspace{0.001cm} 
    
    \includegraphics[width=1.0\linewidth]{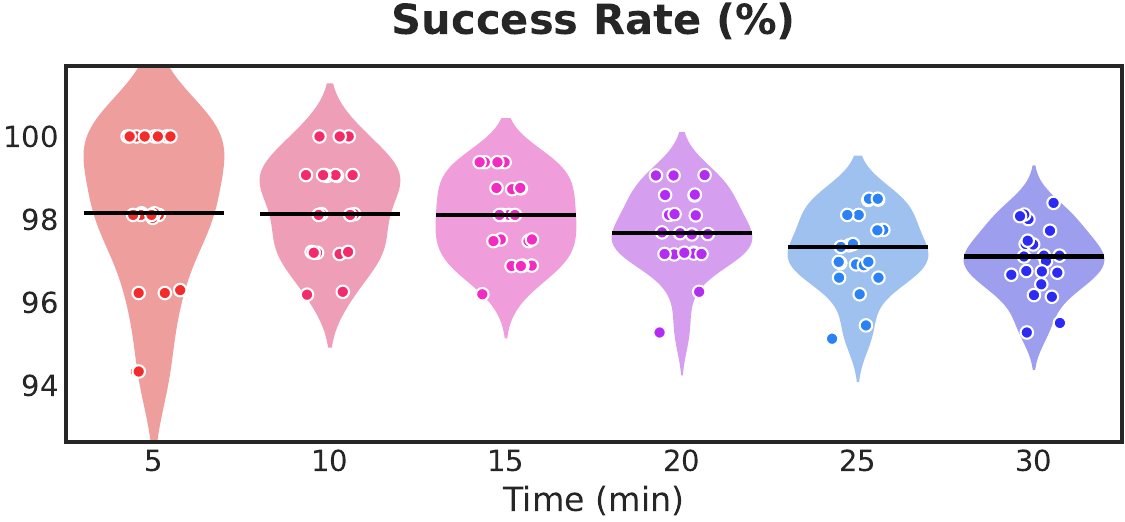}
    
    \vspace{0.001cm} 
    
    \includegraphics[width=1.0\linewidth]{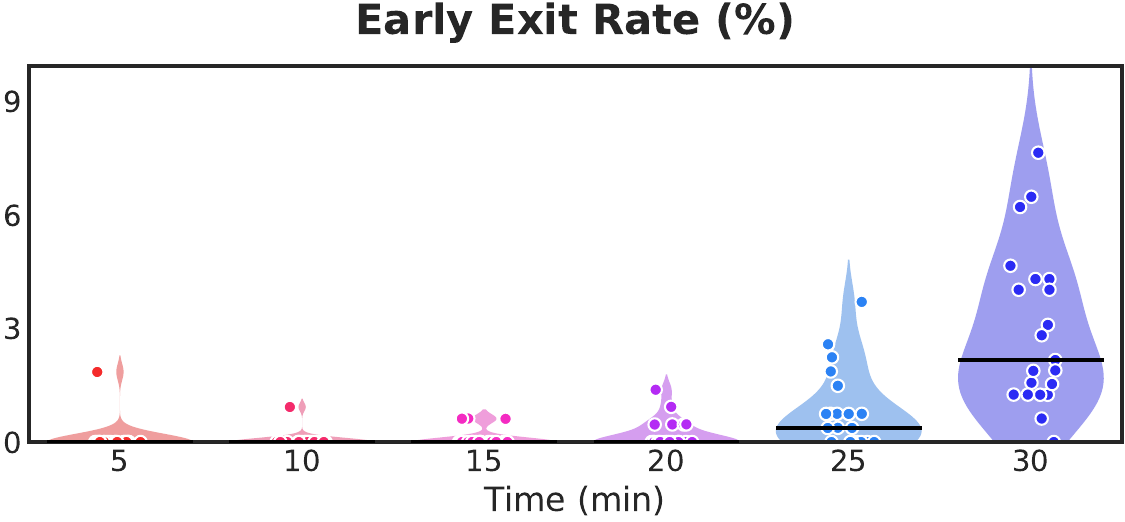}
    
    \caption{\small Experimental results showing the performance of the pipeline, recorded at intervals of five minutes.}
    \label{fig:pipeline_column}
\end{figure}

\noindent Finally, we performed an in-depth analysis of the performance of our system over time, emptying the bin. We conducted 30 independent runs, each lasting 30 min on Object B. Every run began with a bin filled with approximately 350 objects. To~evaluate the pipeline's performance relative to the bin's level, we recorded the system's state every 5~min. The~experimental data are presented in the following figures, organized into three subplots. Each subplot's data are segmented into 5 min intervals, resulting in six violin plots. Each plot visualizes the distribution of the 30 data points (one from each experimental run) and integrates the median~value. Figure~\ref{fig:pipeline_column} shows the high performance of our pipeline, which achieves a median of 600 MPPH and a 97\% success rate even after 30 min. The~plots also clearly illustrate how these metrics change in relation to the bin's fill level. As~mentioned, we started with approximately 350 objects in the bin. A~pick rate of 600 picks per hour means that approximately 300 objects were removed during each run, almost emptying the bin. Both the MPPH and EER plots show that performance worsens over time, particularly in the last 10 min and even more so in the final 5. This is because the objects remaining at the end are usually harder to grasp due to their location near the bin's borders, which are difficult for our gripper to access, due to its size. This difficulty significantly increases the Early Exit Rate and consequently decreases the MPPH. The~success rate also decreases slightly, as~the grasps attempted in these final-stage scenarios are inherently~riskier.

\section{Conclusions}
\noindent Pickalo addresses the challenge of deploying high-throughput and cost-effective bin picking in realistic industrial conditions. The~combination of multi-view scene reconstruction, 6D pose estimation, and~depth enhancement proved to be effective in reconstructing the content of the bin, enabling seamless grasp planning and accurate collision avoidance. Ablation studies confirmed that both the pose buffer module and the depth processing played key roles in the system’s reliability. Across the three evaluated object categories, the~system achieved grasp success rates ranging from $96\%$ to $99\%$, with~approximately 600 mean picks per hour. During~the 30 min bin-emptying experiments, Pickalo maintained a median grasp success rate of $97\%$, demonstrating stable performance as the bin approached an almost empty state. The~ablation studies further confirmed that BridgeDepth and the pose buffer improve both grasp reliability and throughput. The~modularity of the approach allows inspection of partial results of each module, facilitating tracking of potential errors during deployment. Finally, modularity also enables improvement of the pipeline as better foundation models become available.

\bibliographystyle{unsrt}
\bibliography{references}

@ARTICLE{Probabilistic_Multi_View_Fusion,
  author={Yang, Jun and Li, Dong and Waslander, Steven L.},
  journal={IEEE Robotics and Automation Letters}, 
  title={Probabilistic Multi-View Fusion of Active Stereo Depth Maps for Robotic Bin-Picking}, 
  year={2021},
  volume={6},
  number={3},
  pages={4472-4479},
  doi={10.1109/LRA.2021.3068706}}

@article{quaternion_averaging,
  title={Averaging quaternions},
  author={Markley, F Landis and Cheng, Yang and Crassidis, John L and Oshman, Yaakov},
  journal={Journal of Guidance, Control, and Dynamics},
  volume={30},
  number={4},
  pages={1193--1197},
  year={2007}
}

@inproceedings{pitteri2019object,
  title={On object symmetries and 6d pose estimation from images},
  author={Pitteri, Giorgia and Ramamonjisoa, Micha{\"e}l and Ilic, Slobodan and Lepetit, Vincent},
  booktitle={2019 International conference on 3D vision (3DV)},
  pages={614--622},
  year={2019},
  organization={IEEE}
}

@INPROCEEDINGS{MaskRCNN,

  author={He, Kaiming and Gkioxari, Georgia and Dollár, Piotr and Girshick, Ross},

  booktitle={2017 IEEE International Conference on Computer Vision (ICCV)}, 

  title={Mask R-CNN}, 

  year={2017},

  volume={},

  number={},

  pages={2980-2988},

  doi={10.1109/ICCV.2017.322}}

@article{BlenderProc, doi = {10.21105/joss.04901}, url = {https://doi.org/10.21105/joss.04901}, year = {2023}, publisher = {The Open Journal}, volume = {8}, number = {82}, pages = {4901}, author = {Denninger, Maximilian and Winkelbauer, Dominik and Sundermeyer, Martin and Boerdijk, Wout and Knauer, Markus and Strobl, Klaus H. and Humt, Matthias and Triebel, Rudolph}, title = {BlenderProc2: A Procedural Pipeline for Photorealistic Rendering}, journal = {Journal of Open Source Software} }

@inproceedings{bui2024deep6d,
  author    = {Tat Hieu Bui and Yeong Gwang Son and Seung Jae Moon
               and Quang Huy Nguyen and Issac Rhee and Juyong Hong
               and Hyouk Ryeol Choi},
  title     = {Deep Learning Based 6-DoF Antipodal Grasp Planning
               from Point Cloud in Random Bin-Picking Task Using Single-View},
  booktitle = {Proceedings of the {IEEE} International Conference on Robotics and Automation ({ICRA})},
  year      = {2024}
}

@article{fang2023anygrasp,
  author  = {Hao{-}Shu Fang and Chenxi Wang and Hongjie Fang
             and Minghao Gou and Jirong Liu and Hengxu Yan
             and Wenhai Liu and Yichen Xie and Cewu Lu},
  title   = {{AnyGrasp}: Robust and Efficient Grasp Perception
             in Spatial and Temporal Domains},
  journal = {{IEEE} Transactions on Robotics},
  year    = {2023},
  doi     = {10.1109/TRO.2023.3281153}
}

@inproceedings{xiang2018posecnn,
  author    = {Yu Xiang and Tanner Schmidt and Venkatraman Narayanan and Dieter Fox},
  title     = {{PoseCNN}: A Convolutional Neural Network for 6D Object Pose Estimation in Cluttered Scenes},
  booktitle = {Proceedings of Robotics: Science and Systems ({RSS})},
  year      = {2018}
}

@inproceedings{tekin2018realtime,
  author    = {Bugra Tekin and Sudipta N. Sinha and Pascal Fua},
  title     = {Real-Time Seamless Single Shot 6D Object Pose Prediction},
  booktitle = {Proceedings of the {IEEE/CVF} Conference on Computer Vision and Pattern Recognition ({CVPR})},
  pages     = {292--301},
  year      = {2018},
  doi       = {10.1109/CVPR.2018.00038}
}

@inproceedings{wang2019densefusion,
  author    = {Chen Wang and Danfei Xu and Yuke Zhu
               and Roberto Mart{\'{\i}}n{-}Mart{\'{\i}}n
               and Cewu Lu and Li Fei{-}Fei and Silvio Savarese},
  title     = {DenseFusion: 6D Object Pose Estimation by Iterative Dense Fusion},
  booktitle = {Proceedings of the {IEEE/CVF} Conference on Computer Vision and Pattern Recognition ({CVPR})},
  pages     = {3343--3352},
  year      = {2019},
  doi       = {10.1109/CVPR.2019.00345}
}

@misc{Yann2022Megapose,
      title={MegaPose: 6D Pose Estimation of Novel Objects via Render \& Compare}, 
      author={Yann Labbé and Lucas Manuelli and Arsalan Mousavian and Stephen Tyree and Stan Birchfield and Jonathan Tremblay and Justin Carpentier and Mathieu Aubry and Dieter Fox and Josef Sivic},
      year={2022},
      eprint={2212.06870},
      archivePrefix={arXiv},
      primaryClass={cs.CV},
      url={https://arxiv.org/abs/2212.06870}, 
}

@article{li2022s2rpick,
  author  = {Xianzhi Li and Rui Cao and Yidan Feng and Kai Chen
             and Biqi Yang and Chi{-}Wing Fu and Yichuan Li
             and Qi Dou and Yun{-}Hui Liu and Pheng{-}Ann Heng},
  title   = {A Sim-to-Real Object Recognition and Localization Framework
             for Industrial Robotic Bin Picking},
  journal = {{IEEE} Robotics and Automation Letters},
  volume  = {7},
  number  = {2},
  pages   = {3961--3968},
  year    = {2022},
  doi     = {10.1109/LRA.2022.3149026}
}

@inproceedings{labbe2020cosypose,
  author    = {Yann Labb{\'e} and Justin Carpentier and Mathieu Aubry and Josef Sivic},
  title     = {{CosyPose}: Consistent Multi-view Multi-object 6D Pose Estimation},
  booktitle = {Proceedings of the European Conference on Computer Vision ({ECCV})},
  year      = {2020}
}

@InProceedings{Huang_2025_ICCV,
    author    = {Huang, Ziqin and Li, Chengxi and Li, Yingyue and Liu, Xingyu and Zhang, Chenyangguang and Zhang, Ruida and Fu, Bowen and Hu, Xinggang and Qu, Yun and Liu, Mengge and Mao, Yixiu and Huang, Wendong and Wang, Gu and Ji, Xiangyang},
    title     = {Lessons and Winning Solutions in Industrial Object Detection and Pose Estimation from the 2025 Bin-Picking Perception Challenge},
    booktitle = {Proceedings of the IEEE/CVF International Conference on Computer Vision (ICCV) Workshops},
    month     = {October},
    year      = {2025},
    pages     = {2408-2414}
}

@inproceedings{fu2024lowcost,
  author    = {Xingdou Fu and Lin Miao and Yasuhiro Ohnishi
               and Yuki Hasegawa and Masaki Suwa},
  title     = {A Low-Cost, High-Speed, and Robust Bin Picking System for Factory Automation
               Enabled by a Non-Stop, Multi-View, and Active Vision Scheme},
  booktitle = {Proceedings of the {IEEE/RSJ} International Conference on Intelligent Robots and Systems ({IROS})},
  year      = {2024},
  pages     = {11566--11573}
}

@inproceedings{guan2025bridgedepth,
  author    = {Tongfan Guan and Jiaxin Guo and Chen Wang and Yun{-}Hui Liu},
  title     = {{BridgeDepth}: Bridging Monocular and Stereo Reasoning with Latent Alignment},
  booktitle = {Proceedings of the {IEEE/CVF} International Conference on Computer Vision ({ICCV})},
  year      = {2025},
  pages     = {27681--27691},
  note      = {Highlight paper},
  url       = {https://arxiv.org/abs/2508.04611}
}

@article{wen2025foundstereo,
  title={FoundationStereo: Zero-Shot Stereo Matching},
  author={Bowen Wen and Matthew Trepte and Joseph Aribido and Jan Kautz and Orazio Gallo and Stan Birchfield},
  journal={CVPR},
  year={2025}
}

@INPROCEEDINGS{Rodrigues2012,
  author={Rodrigues, José Jeronimo and Kim, Jun-Sik and Furukawa, Makoto and Xavier, João and Aguiar, Pedro and Kanade, Takeo},
  booktitle={2012 IEEE/RSJ International Conference on Intelligent Robots and Systems}, 
  title={6D pose estimation of textureless shiny objects using random ferns for bin-picking}, 
  year={2012},
  volume={},
  number={},
  pages={3334-3341},
  doi={10.1109/IROS.2012.6385680}}

@INPROCEEDINGS{CAD_based_recognition,
  author={Ulrich, Markus and Wiedemann, Christian and Steger, Carsten},
  booktitle={2009 IEEE International Conference on Robotics and Automation}, 
  title={CAD-based recognition of 3D objects in monocular images}, 
  year={2009},
  volume={},
  number={},
  pages={1191-1198},
  doi={10.1109/ROBOT.2009.5152511}}

@InProceedings{Drost2010,
  author    = {Bertram Drost and Markus Ulrich and Nassir Navab and Slobodan Ili{\'c}},
  title     = {Model globally, match locally: Efficient and robust 3D object recognition},
  booktitle = {Proceedings of the IEEE Conference on Computer Vision and Pattern Recognition (CVPR)},
  pages     = {998--1005},
  year      = {2010}
}

@INPROCEEDINGS{Cad_based_pose,
  author={Wu, Cheng-Hei and Jiang, Sin-Yi and Song, Kai-Tai},
  booktitle={2015 15th International Conference on Control, Automation and Systems (ICCAS)}, 
  title={CAD-based pose estimation for random bin-picking of multiple objects using a RGB-D camera}, 
  year={2015},
  volume={},
  number={},
  pages={1645-1649},
  doi={10.1109/ICCAS.2015.7364621}}

@INPROCEEDINGS{VincentLepetit,
  author={Hinterstoisser, Stefan and Holzer, Stefan and Cagniart, Cedric and Ilic, Slobodan and Konolige, Kurt and Navab, Nassir and Lepetit, Vincent},
  booktitle={2011 International Conference on Computer Vision}, 
  title={Multimodal templates for real-time detection of texture-less objects in heavily cluttered scenes}, 
  year={2011},
  volume={},
  number={},
  pages={858-865},
  doi={10.1109/ICCV.2011.6126326}}

@article{Region-Aware,
author = {Zhong, Xungao and Gong, Tao and Yu, Junzhi and Luo, Jiaguo and Zhou, Chengxian and Zhong, Xunyu and Liu, Qiang},
year = {2025},
month = {01},
pages = {1-1},
title = {Region-Aware Grasping for Stacked Workpieces: A 6D-Wise Label Self-Generation Method and Robust Evaluation Strategy},
volume = {PP},
journal = {IEEE Transactions on Automation Science and Engineering},
doi = {10.1109/TASE.2025.3580784}
}

@article{Lin2023SAM6D,
  author  = {Jiehong Lin and Lihua Liu and Dekun Lu and Kui Jia},
  title   = {{SAM-6D}: Segment Anything Model Meets Zero-Shot 6D Object Pose Estimation},
  journal = {arXiv preprint arXiv:2311.15707},
  year    = {2023}
}

@article{Sun2024,
  author  = {Han Sun and Zhuangzhuang Zhang and Haili Wang and Yizhao Wang and Qixin Cao},
  title   = {A Novel Robotic Grasp Detection Framework Using Low-Cost RGB-D Camera for Industrial Bin Picking},
  journal = {IEEE Trans. on Instrumentation and Measurement},
  volume  = {73},
  pages   = {1--12},
  year    = {2024}
}

@misc{Caraffa2025,
      title={Accurate and efficient zero-shot 6D pose estimation with frozen foundation models}, 
      author={Andrea Caraffa and Davide Boscaini and Fabio Poiesi},
      year={2025},
      eprint={2506.09784},
      archivePrefix={arXiv},
      primaryClass={cs.CV},
      url={https://arxiv.org/abs/2506.09784}, 
}

@misc{wen2024foundationposeunified6dpose,
      title={FoundationPose: Unified 6D Pose Estimation and Tracking of Novel Objects}, 
      author={Bowen Wen and Wei Yang and Jan Kautz and Stan Birchfield},
      year={2024},
      eprint={2312.08344},
      archivePrefix={arXiv},
      primaryClass={cs.CV},
      url={https://arxiv.org/abs/2312.08344}, 
}

@article{Peiyuan,
   title={Reasoning and Learning a Perceptual Metric for Self-Training of Reflective Objects in Bin-Picking With a Low-Cost Camera},
   volume={10},
   ISSN={2377-3774},
   url={http://dx.doi.org/10.1109/LRA.2025.3600139},
   DOI={10.1109/lra.2025.3600139},
   number={10},
   journal={IEEE Robotics and Automation Letters},
   publisher={Institute of Electrical and Electronics Engineers (IEEE)},
   author={Ni, Peiyuan and Chew, Chee Meng and Ang, Marcelo H. and Chirikjian, Gregory S.},
   year={2025},
   month=oct, pages={10458–10465} }

@misc{huang2025xyzibdhighprecisionbinpickingdataset,
      title={XYZ-IBD: A High-precision Bin-picking Dataset for Object 6D Pose Estimation Capturing Real-world Industrial Complexity}, 
      author={Junwen Huang and Jizhong Liang and Jiaqi Hu and Martin Sundermeyer and Peter KT Yu and Nassir Navab and Benjamin Busam},
      year={2025},
      eprint={2506.00599},
      archivePrefix={arXiv},
      primaryClass={cs.CV},
      url={https://arxiv.org/abs/2506.00599}, 
}

@misc{SAM,
      title={Segment Anything}, 
      author={Alexander Kirillov and Eric Mintun and Nikhila Ravi and Hanzi Mao and Chloe Rolland and Laura Gustafson and Tete Xiao and Spencer Whitehead and Alexander C. Berg and Wan-Yen Lo and Piotr Dollár and Ross Girshick},
      year={2023},
      eprint={2304.02643},
      archivePrefix={arXiv},
      primaryClass={cs.CV},
      url={https://arxiv.org/abs/2304.02643}, 
}

@INPROCEEDINGS {Defomo,
author = { Jiang, Hualie and Lou, Zhiqiang and Ding, Laiyan and Xu, Rui and Tan, Minglang and Jiang, Wenjie and Huang, Rui },
booktitle = { 2025 IEEE/CVF Conference on Computer Vision and Pattern Recognition (CVPR) },
title = {{ DEFOM-Stereo: Depth Foundation Model Based Stereo Matching }},
year = {2025},
volume = {},
ISSN = {},
pages = {21857-21867},
doi = {10.1109/CVPR52734.2025.02036},
url = {https://doi.ieeecomputersociety.org/10.1109/CVPR52734.2025.02036},
publisher = {IEEE Computer Society},
address = {Los Alamitos, CA, USA},
month =Jun}

@INPROCEEDINGS{Robi,
  author={Yang, Jun and Gao, Yizhou and Li, Dong and Waslander, Steven L.},
  booktitle={2021 IEEE/RSJ International Conference on Intelligent Robots and Systems (IROS)}, 
  title={ROBI: A Multi-View Dataset for Reflective Objects in Robotic Bin-Picking}, 
  year={2021},
  volume={},
  number={},
  pages={9788-9795},
  doi={10.1109/IROS51168.2021.9635871}}

@misc{kleeberger2021automaticgraspposegeneration,
      title={Automatic Grasp Pose Generation for Parallel Jaw Grippers}, 
      author={Kilian Kleeberger and Florian Roth and Richard Bormann and Marco F. Huber},
      year={2021},
      eprint={2104.11660},
      archivePrefix={arXiv},
      primaryClass={cs.RO},
      url={https://arxiv.org/abs/2104.11660}, 
}

\begin{IEEEbiography}
[{\includegraphics[width=1in,height=1.25in,clip,keepaspectratio]{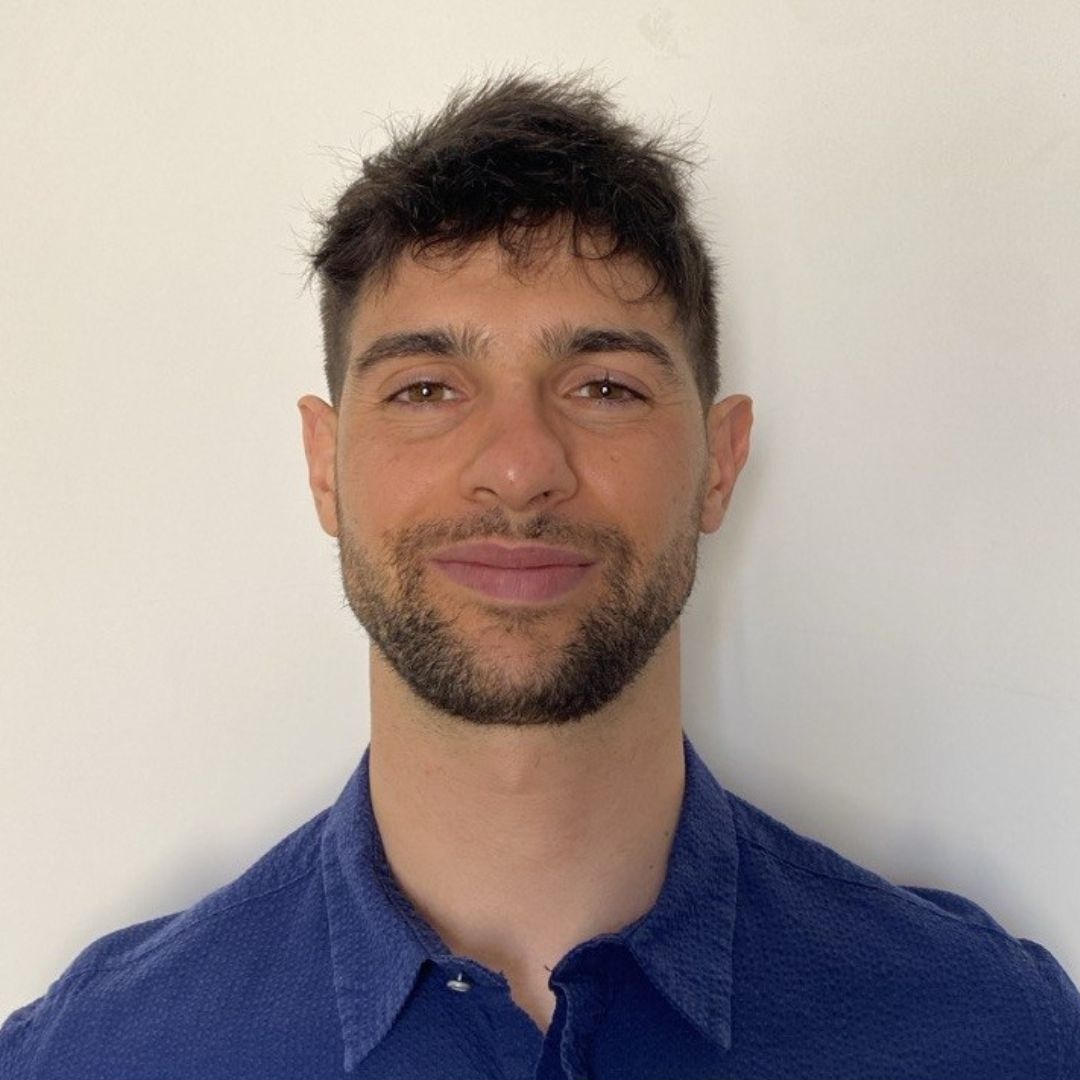}}]{Alessandro Tarsi} received the B.S. and M.S. degrees in Automation Engineering from the University Politecnica delle Marche, and the University of Bologna, respectively. He worked as a Robotics Engineer for the MESH facility (formerly iCub Tech) at the Istituto Italiano di Tecnologia. He currently works as a Research Engineer at the Institut des Systèmes Intelligents et de Robotique (Sorbonne University), Paris, France.
\end{IEEEbiography}
\vspace{-15pt}
\begin{IEEEbiography}
[{\includegraphics[width=1in,height=1.25in,clip,keepaspectratio]{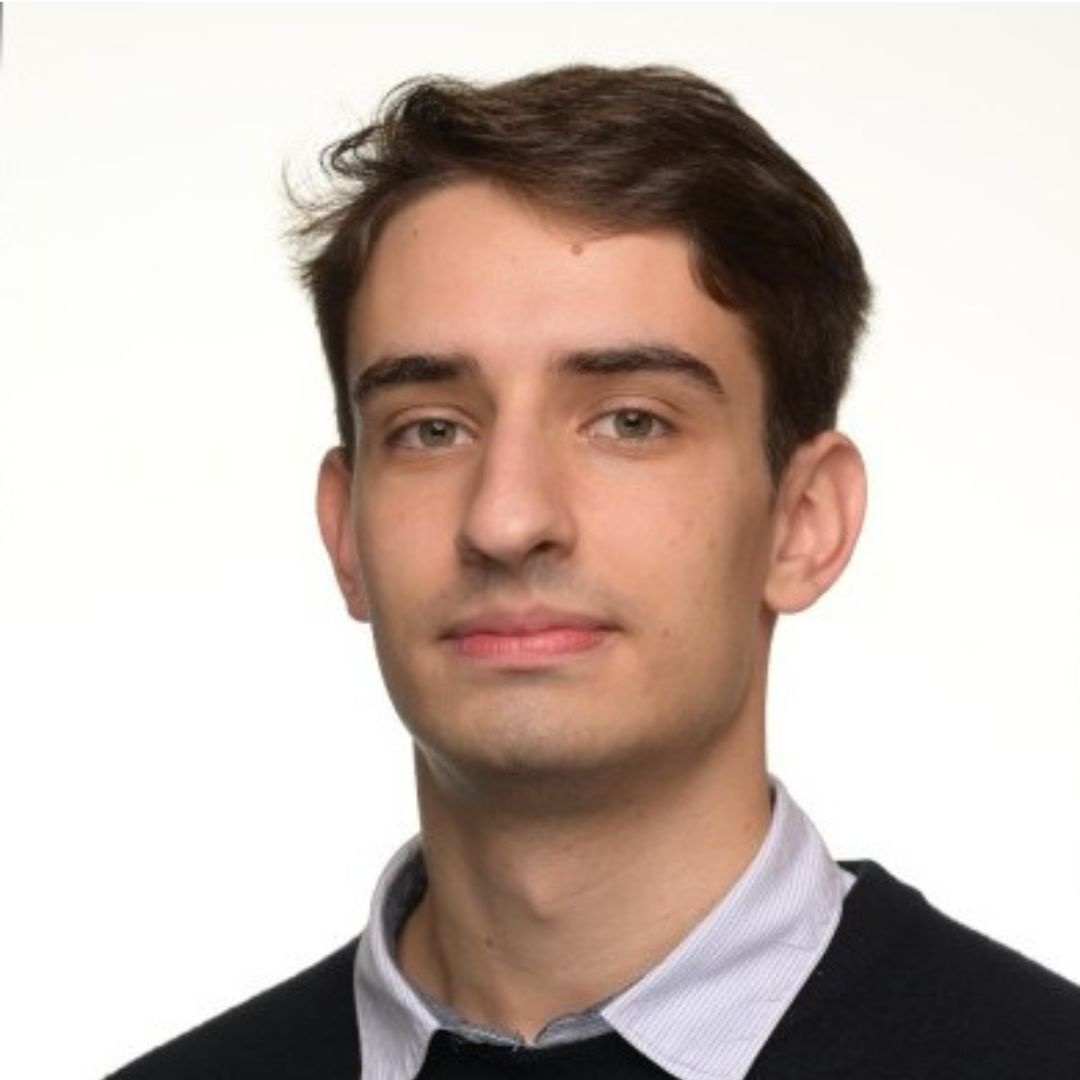}}]{Matteo Mastrogiuseppe} received the B.Sc. in Mechanical Engineering from Politecnico di Torino. He later graduated in Autonomous Systems in a M.Sc. Double Degree Program from University of Trento and Aalto University. He worked as a Robotics Engineer for the MESH facility (formerly iCub Tech) at the Istituto Italiano di Tecnologia. He currently works as a Robotics Engineer at Generative Bionics in Genoa, Italy.
\end{IEEEbiography}
\vspace{-15pt}
\begin{IEEEbiography}
[{\includegraphics[width=1in,height=1.25in,clip,keepaspectratio]{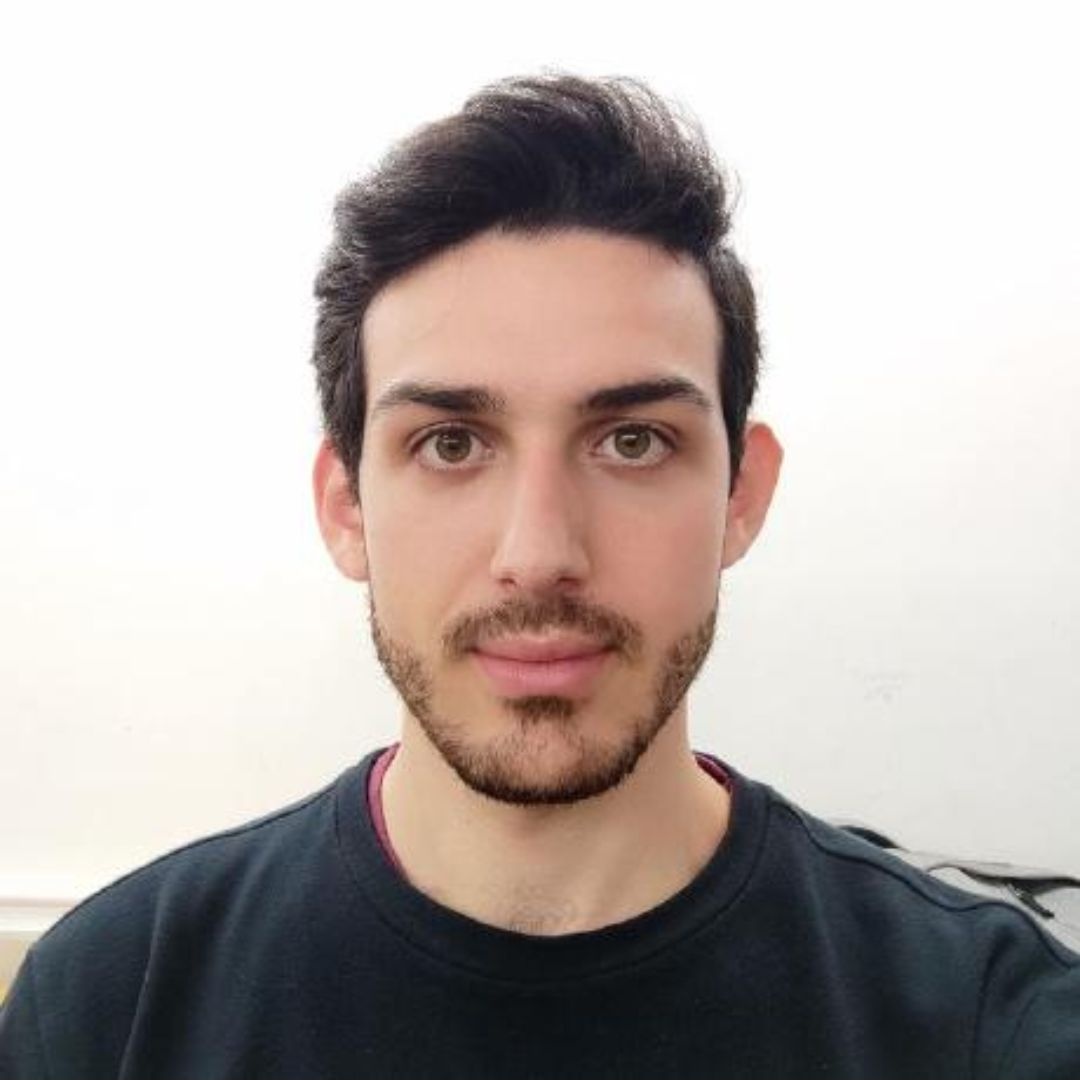}}]{Saverio Taliani} received his B.S. in Automation and Computer Engineering and his M.S. in Control Engineering from the Sapienza University of Rome. Previously, he worked as a Research Fellow in the Artifical and Mechanical Intelligence research line at the Istituto Italiano di Tecnologia. He currently works as a Robotics Engineer at Generative Bionics in Genoa, Italy.
\end{IEEEbiography}
\vspace{-15pt}
\begin{IEEEbiography}
[{\includegraphics[width=1in,height=1.25in,clip,keepaspectratio]{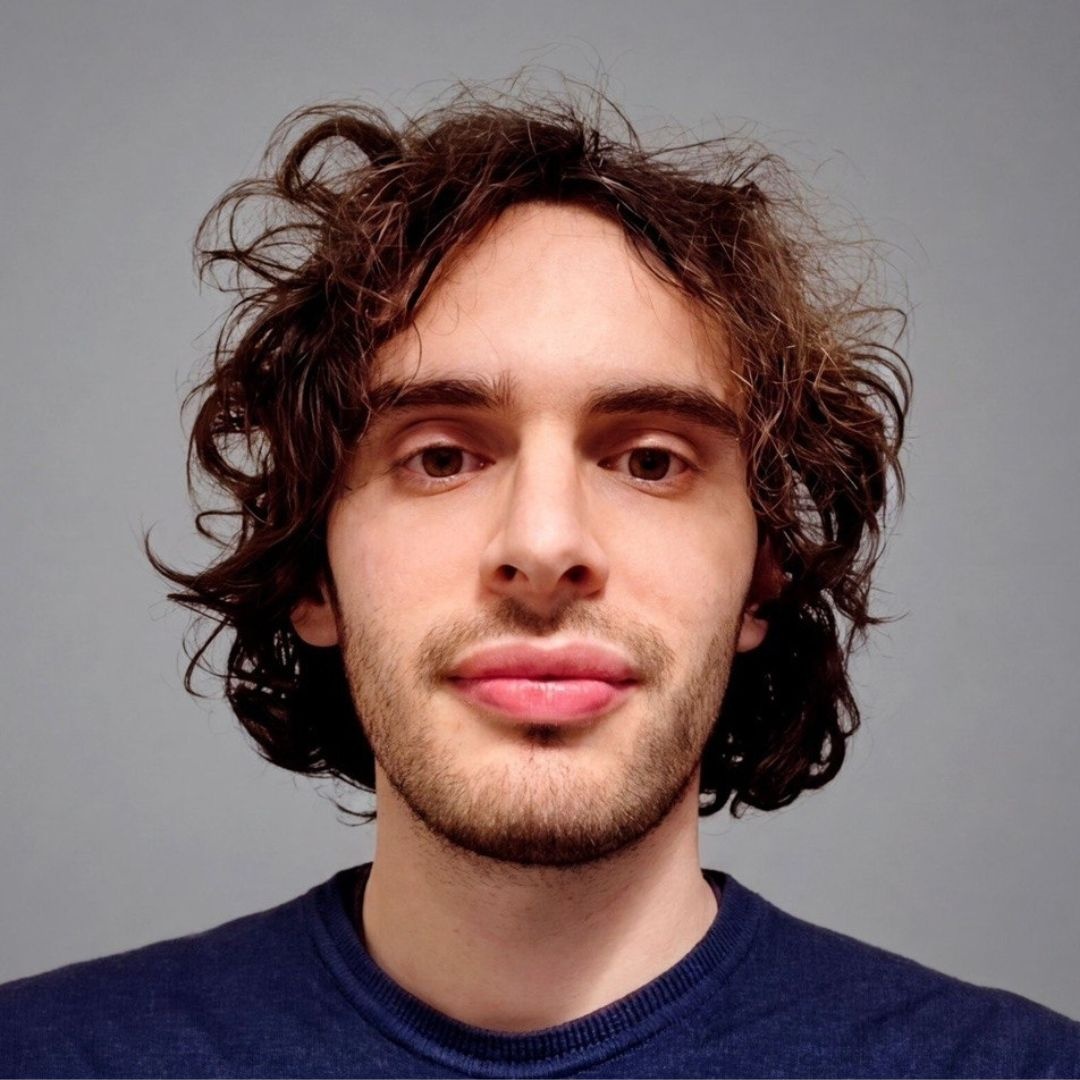}}]{Simone Cortinovis} received his B.S. and M.S. degrees in Mechatronics Engineering from the University of Bergamo, Italy. Previously, he worked as a Robotics Engineer with the MESH facility (formerly iCub Tech) at the Istituto Italiano di Tecnologia. He is currently a Robotics Engineer at Generative Bionics in Genoa, Italy.
\end{IEEEbiography}
\vspace{-15pt}
\begin{IEEEbiography}
[{\includegraphics[width=1in,height=1.25in,clip,keepaspectratio]{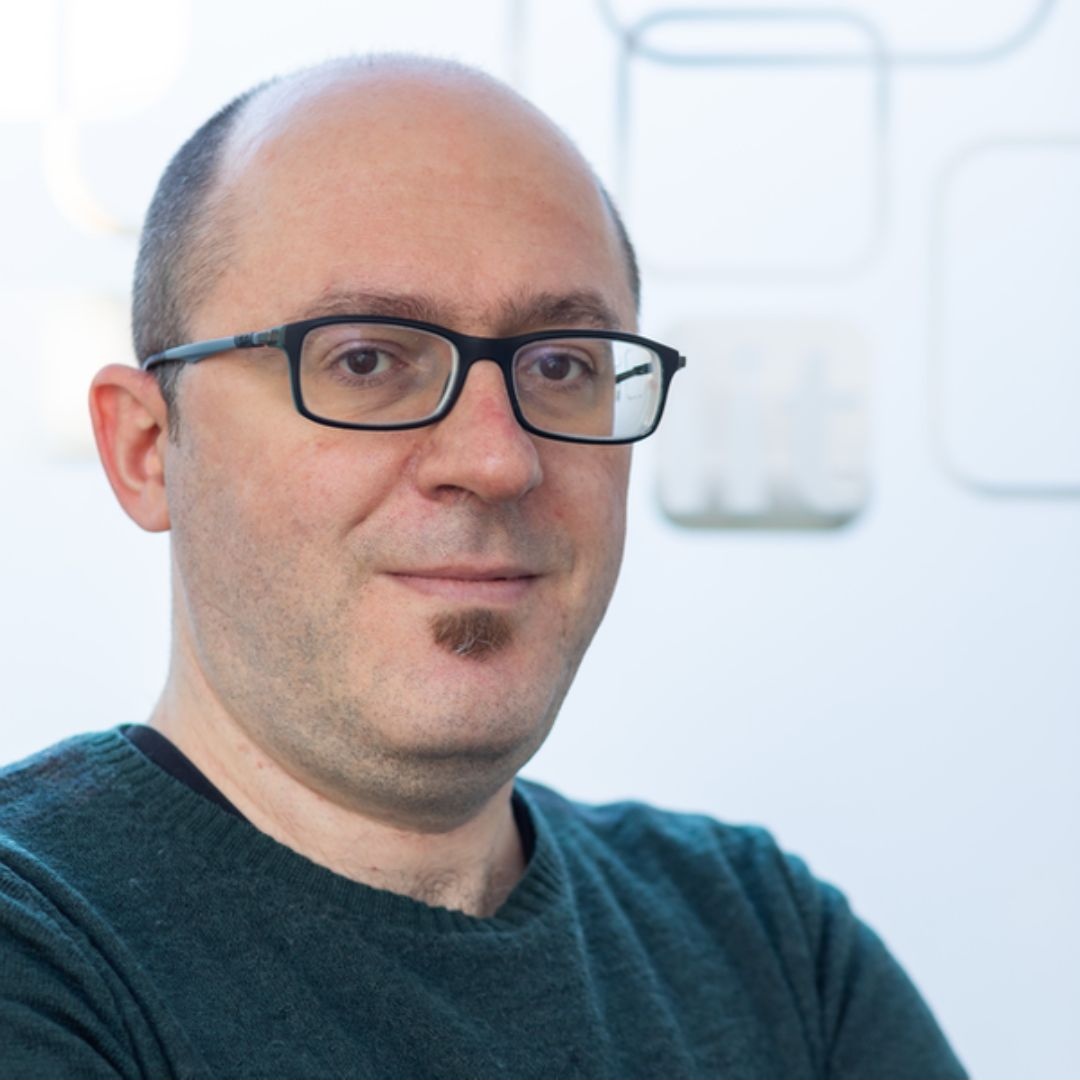}}]{Ugo Pattacini} Dr. Ugo Pattacini is a Technologist coordinating the MESH facility at IIT, where he oversees the optimization and mechatronic evolution of the iCub and R1 humanoids. In the past, he gathered deep working experience in hi-tech companies in Formula 1 applications (Magneti Marelli Racing Dep. and Toyota F1 Team) and aerospace (Thales Alenia Space). He holds a Ph.D. in Robotics from IIT and since 2008 has been involved in the development of humanoid platforms through several EU projects (e.g., RobotCub, CHRIS, EFAA, WYSIWYD, TACMAN, ETAPAS), focusing his interests on the advancement of motor capabilities. He is a senior member of RAS and CSS societies of IEEE.
\end{IEEEbiography}

\end{document}